\documentclass[11pt]{article}                 
\usepackage{graphicx}
 \usepackage[numbers,sort]{natbib}
\usepackage{shortcuts}
\usepackage{tikz}
\usetikzlibrary{bayesnet}
\usetikzlibrary{arrows}
\usetikzlibrary{fit}
\usepackage{color}
\usepackage{amsmath,amssymb}
\usepackage{booktabs}
\usepackage{algorithm}
\usepackage{algorithmic}
\usepackage{shortcuts}
\usepackage{bbm}
\usepackage{comment}
\usepackage{xcolor}
\usepackage{array}
\usepackage{makecell}
\usepackage{soul}
\usepackage{url}
\usepackage{mwe}
  
\usepackage{xcolor}
\usetikzlibrary{calc}
\usepackage{bbm}
\usepackage{subfig}
\usepackage{adjustbox}
\usetikzlibrary{shapes,arrows}
\newtheorem{theorem}{Theorem}
\newtheorem{definition}{Definition}
\newtheorem{lemma}{Lemma}

\newenvironment{ppl}{\fontfamily{cmtt}\selectfont}{\par}

\usetikzlibrary{decorations.pathreplacing,automata,calc,positioning}
\tikzset{
  latentnode/.style={draw, minimum width=5mm, shape=circle, ultra thick, black},
  dagconn/.style={arrows=->, black, thick},
  plate/.style={draw, shape=rectangle, rounded corners=0.5ex, thick,
    minimum width=3.1cm, text width=3.1cm, align=right, inner sep=10pt, inner ysep=10pt,label={[xshift=-14pt,yshift=14pt]south east:#1}}
}

\graphicspath{{figures/}}

\usepackage{xcolor}

\begin{document}

\title{IntelligentPooling:
\\Practical Thompson Sampling for mHealth 
}

\author{Sabina Tomkins   \\ Stanford University  \\ stomkins@stanford.edu\\ \and     
        Peng Liao \\ Harvard University \\ pengliao@g.harvard.edu \and
        Predrag Klasnja  \\ University of Michigan \\ klasnja@umich.edu\and
        Susan Murphy  \\ Harvard University \\ samurphy@fas.harvard.edu
}

\maketitle

\begin{abstract}
In mobile health (mHealth) smart devices deliver behavioral treatments repeatedly over time to a user with the goal of helping the user adopt and maintain  healthy behaviors. Reinforcement learning appears ideal for learning how to optimally make these sequential treatment decisions. 
However, significant challenges must be overcome before reinforcement learning 
can be effectively deployed in a mobile  healthcare setting.
In this work we are concerned with the following challenges:
1) individuals who are in the same context  can exhibit differential response to treatments   
2) only a limited amount of  data is available for learning on any one individual,  and 
3) non-stationary responses to treatment.
To address these challenges we generalize  Thompson-Sampling bandit algorithms to develop  \ourapproach{}. 
\ourapproach{}  learns personalized treatment policies 
thus addressing challenge one.
To address the second challenge, \ourapproach{} updates each user's degree of personalization
while making use of available data on other users to speed up learning. 
Lastly, \ourapproach{} allows responsivity to vary as a function of a user's time since beginning treatment, thus addressing challenge three. 

\end{abstract}

\section{Introduction}
\label{sec:intro}

Mobile health (mHealth) applications deliver  treatments in users' everyday lives
to support  healthy behaviors.
These mHealth applications offer an opportunity 
to impact health across  a diverse range of domains
from substance use \cite{rabbi2017sara}, to disease self-management \cite{hamine2015impact} to physical inactivity \cite{consolvo2008activity}.
For example, to help users increase their physical activity, an mHealth application
might send walking suggestions at the  times and in the contexts (e.g. current location or recent physical activity) 
when a user is likely to be able  
to pursue the suggestions. 
A goal of mHealth applications is to 
provide treatments in contexts in which users need support \textit{while} 
avoiding over-treatment. 
Over-treatment can lead to user disengagement \cite{nahum2017just}, for example 
users might  ignore treatments or even delete the application. 
 Consequently, the  goal 
 is to be able to learn an optimal policy for when and how to intervene for each user and context
 without over-treating.

 Contextual bandit algorithms appear ideal for this task.
 Contextual bandit algorithms have been successful in a range of application settings from news recommendations \cite{li2010contextual} to education \cite{NEURIPS2018_d8c9d05e}.
 However, as we discuss below, many challenges remain to adapt contextual bandit algorithms for mHealth settings.
 Thompson sampling  offers an attractive framework for addressing these challenges.
 In their seminal work \cite{agrawal2013thompson}, Agrawal and Goyal show that 
 Thompson sampling for contextual bandits, which works well in practice, can also achieve strong theoretical guarantees.
 In our work, we propose Thompson sampling contextual bandit algorithm which 
 introduces a mixed effects structure for the weights on the feature vector, an algorithm 
 we call \ourapproach. 
 We demonstrate empirically that \ourapproach{} has many advantages. 
We also derive a 
 high-probability regret bound for our approach which achieves similar regret to \cite{agrawal2013thompson}.  
Unlike \cite{agrawal2013thompson}, our regret bound  depends on the variance components
introduced by the mixed effects structure which 
 is at the center of our approach.

\subsection{Challenges}
There are significant challenges to learning optimal policies in  mHealth.
This work primarily addresses the challenge of learning personalized user policies from limited data. 
Contextual bandit algorithms can be viewed as algorithms that use the user's context to \textit{adapt}  treatment. While this approach can have advantages compared to ignoring the user's context, it fails to address that
users can respond differentially to treatments even when they appear to be in the same context. This  occurs since sensors on smart devices are unlikely to record all aspects of a user's context that affect their health behaviors.   For example, the context may not include social constraints on the user (e.g.,  care-giving responsibilities), which may  influence  the user's ability to be active. Thus, algorithms that can learn from the differential responsiveness to treatment are desirable.
This motivates the need for an  algorithm that not only incorporates contextual information, but that can also learn personalized policies. A natural first approach would be to use the  algorithm separately for each user, but the  algorithm is likely to learn very slowly if data on a user is sparse and/or noisy.  However, typically in mHealth studies multiple users are using the application at any given time.   Thus an  algorithm that pools data over users intelligently so as to speed up learning of personalized policies is desirable.

An additional challenge  is non-stationary responses to treatment (e.g. non-stationary reward function). 
For example,
in the beginning of a study, a user might be excited to receive a treatment, however after a few weeks this excitement can wane. This motivates the need for 
algorithms that can learn time-varying treatment policies. 

\subsection{Contributions}
We develop \ourapproach{}, a type of  Thompson sampling contextual bandit algorithm specifically designed to overcome the above challenges. 
Our main contributions are:

\begin{itemize}
\item[--] \ourapproach{}:  \textit{A Thompson sampling contextual bandit algorithm for rapid personalization in limited data settings}. 
This algorithm employs classical random effects in the reward function \cite{raudenbush2002hierarchical,laird1982random} and empirical Bayes \cite{morris1983parametric,casella1985introduction}) to adaptively adjust the degree to which policies are personalized to each user.
We present an analysis of this adaptivity in \secref{subsec:intuition} showing that \ourapproach{} can learn to personalize to a user
as a function of the observed variance in the treatment effect both between and within users. 
\item[--] A high probability regret bound for \ourapproach. 
\item[--] \textit{An empirical evaluation of \ourapproach{} in a simulation environment constructed from  mHealth data}. \ourapproach{}
not only achieves 26\% lower regret than state-of-the-art approaches, it also is better able to adapt to the degree of heterogeneity present in a population
than this approach. 
\item[--] \textit{ Feasibility of \ourapproach{} from a pilot study in a live clinical trial}. We demonstrate that \ourapproach{} can be executed in a real-time online environment
and show preliminary evidence of this method's effectiveness. 
\item[--] We show how to modify \ourapproach{} to learn in non-stationary environments. 
\end{itemize}

 Next, in \secref{sec:related_work} we discuss relevant related work. In \secref{sec:approach} we present \ourapproach{} and provide a high-probability regret bound for this algorithm.
 We then describe how we use historical data to construct a simulation environment and evaluate our approach against state-of-the-art in \secref{sec:experimental_design}. Next, in \secref{sec:clinical} we introduce the feasibility study and provide preliminary evidence into the benefits of this approach. We then discuss how to extend this work to include time-varying effects in \secref{sec:time}. Finally, we discuss the limitations with our approach in \secref{sec:limitations} before concluding.

\section{Related Work}
\label{sec:related_work}
To put the proposed work in a broader healthcare perspective, an overview of similar work in mHealth is provided by \secref{sec:rl_mhealth}. Next, we discuss the
extent to which reinforcement learning/bandit algorithms have been deployed in mHealth settings (\secref{sec:rl_mhealth}). \ourapproach{} has similarities with several modeling approaches, here 
we discuss the most relevant:
 multi-task learning, meta-learning, Gaussian processes for Thompson Sampling contextual bandits, and time-delayed bandits. These topics are discussed in \secref{sec:rel_mt} - \secref{sec:rel_time}.

\subsection{Connections to Bandit algorithms in mHealth}
\label{sec:rl_mhealth}
Bandit algorithms in mHealth have 
 typically used one of two approaches. The  first approach is person specific, that is,  an algorithm is deployed separately on each user, such as in \cite{rabbi2015mybehavior}, \cite{jaimes2016preventer},  \cite{forman2018can} and \cite{liao2020personalized}. This approach makes sense when users are highly heterogeneous, that is, their optimal policies differ greatly one from another.   However, this approach can present challenges for policy learning  when data is scarce and/or noisy, as in our motivating example of encouraging activity in an mHealth study where only a few decision time-points occur each day (see \citet{xia2018price} for an empirical evaluation of the 
shortcomings of Thompson sampling for personalized contextual bandits in mHealth settings). The second approach completely pools users' data, that is one  algorithm is used on all users so as to learn a common  treatment policy
both in bandit algorithms  \cite{paredes2014poptherapy,yom2017encouraging}, and in full reinforcement learning algorithms \cite{clarke2017mstress,zhou2018personalizing}. This second approach can potentially learn quickly but may result in poor performance if there is large heterogeneity between users. 
We compare to these two approaches empirically as they not only represent state-of-the-art in 
practice, they also represent two intuitive theoretical extremes.

In \ourapproach{} we strike a balance between these two extremes, adjusting the degree of pooling  to the degree that users are similarly responsive.
When users are heterogeneous, \ourapproach{} achieves lower regret than the second approach while learning more quickly than the first approach. When users are homogeneous our method performs as well as the second approach.

\subsection{Connections to multi-task learning and meta-learning}
\label{sec:rel_mt}
Following original work on non-pooled linear contextual bandits\cite{agrawal2013thompson}, researchers have proposed pooling  data in a variety of ways. For example, Deshmukh et al. \cite{deshmukh2017multi} proposed pooling data from different arms of a single bandit problem. Li and Kar \cite{li2015context} used context-sensitive clustering to produce aggregate reward estimates for the bandit algorithm.  More relevant to this work is multi-task Gaussian Process (GP), e.g., \cite{lawrence2004learning,bonilla2008multi,wang2012nonparametric}, however these have been proposed in the prediction as opposed to the reinforcement learning setting. The Gang of Bandits approach \cite{cesa2013gang}, which is a generalization from the original LinUCB algorithm for a single task \cite{li2010contextual}, has been shown to be successful when there is prior knowledge on the similarities between users. For example, a known social network graph might provide a mechanism for pooling. It was later extended to the Horde of Bandits in \cite{vaswani2017horde} which used Thompson Sampling, allowing the algorithm to deal with a large number of tasks. 

Each of the multi-task approaches introduces some concept of similarity between users. The extent to which a given user's data contributes to another user's policy is some function of this similarity measure. This is fundamentally different from the approach taken in \ourapproach{}.  Rather than determining the extent to which any two users are similar, \ourapproach{} determines the extent to which a given user's reward function parameters differ from parameters in a population (average over all users)  reward function.
This approach has the advantage of requiring fewer hyper-parameters, as we do not need to learn a similarity function between users. Instead of a pairwise similarity function it is as if we are learning a similarity between each user and the population average. In the limited data setting, we expect this simpler model to be advantageous.

In meta-learning, one exploits shared structure across tasks to improve performance on new tasks. \ourapproach{} thus shares similarities with  meta-learning for reinforcement learning \cite{nagabandi2018deep,finn2019online,finn2018probabilistic,zintgraf2019caml,gupta2018meta,saemundsson2018meta}. At a high level, one can view our method  as a form of meta-learning where the population-level parameters are learned from all available data and each user's parameters represent deviations from the shared parameters. However, 
 while meta-learning might require a large collection of source tasks, we demonstrate the efficacy of our approach on data on the  small scale found in clinical mHealth studies. 
 
\subsection{Connections to Gaussian process models for Thompson sampling contextual bandits}
\label{sec:rel_gp}
\ourapproach{} is based on Bayesian mixed effects model of the reward, which is similar to using a Gaussian Process (GP) model with a simple form of the kernel. GP models have been used for multi-armed bandits \cite{chowdhury2017kernelized,brochu2010portfolio,srinivas2009gaussian,desautels2014parallelizing,wang2016optimization,djolonga2013high,bogunovic2016time} , and  for contextual bandits  \cite{li2010contextual,krause2011contextual}.  However the above approaches do not structure the way in which the pooling of data across users occurs.   \ourapproach{} uses a  mixed effects GP model to pool across users in structured manner.   Although mixed effects GP models have been previously used for off-line data analysis \cite{shi2012mixed,luo2018mixed}, to the best of our knowledge they have not been previously used in the online decision making setting considered in this work.

 \subsection{Connection to non-stationary linear bandits}
 \label{sec:rel_time}
There is a growing literature  investigating how to adapt linear bandit algorithms to changing environments. A common approach is for the learning algorithm to differentially weight data across time. 
Differential weighting is used by both  \citet{russac2019weighted} (using a  LinUCB algorithm) and \citet{kim2019near} (using  perturbation-based algorithms).   \citet{cheung2018learning} use a linear  moving window to estimate the parameters in the reward function and \citet{zhao2020} restart the algorithm at regular intervals discarding the prior data.  Similarly \citet{bogunovic2016time}, using  GP-based UCB algorithms,   accommodate non-stationarity by both  restarting and using an autoregressive model for the rewards function.  \citet{kim2020randomized} analyze the non-stationary setting 
with randomized exploration.

\ourapproach{} allows for   non-stationary reward functions  by the use of time-varying random effects. The correlation between the time-varying random effects induces a weighted estimator whereby more weight is put on the recently collected samples, similar to the discounted estimators in \cite{russac2019weighted} and \cite{kim2019near}. In contrast to existing approaches, \ourapproach{} considers both individual and time-specific variation.

\section{Intelligent Pooling}
\label{sec:approach}
\ourapproach{}  is a generalization of  a Thompson sampling contextual bandit for learning personalized treatment policies. 
We first outline the components of  \ourapproach{} 
 and then introduce the problem definition in \secref{sec: notation}. As our approach offers a natural alternative to 
 two commonly used approaches, we begin by describing these simpler methods in \secref{sec:bandit_formulation}.
 We introduce our method in \secref{sec:pooling_method}.

 \subsection{Overview}
 \label{sec: overview}
The central component of \ourapproach{} is a Bayesian  model for the reward function. In particular,  \ourapproach{} uses a Gaussian mixed effects linear model for the reward function.  Mixed effects models are widely used across the health and behavioral sciences to model the variation in the linear model parameters across users \cite{raudenbush2002hierarchical,laird1982random} and within a user across time. Use of these  models enhances the ability of  domain scientists  to inform and critique the model used in \ourapproach{}.   
 The properties and 
 pitfalls of these models are well understood; see \cite{qian2019linear} for an application of a mixed effects model in mHealth.   
\ourapproach{} uses Bayesian inference for the mixed effects model. As discussed in Section \ref{sec:rel_gp}, a Bayesian mixed effects linear model is a GP model with a simple kernel.  This facilitates increasing the flexibility of the model for the reward function, given sufficient data.

Furthermore, \ourapproach{} uses Thompson sampling \cite{thompson1933likelihood}, also known as posterior sampling \cite{russo2014learning}, to select actions.
At each decision point, the parameters in the model for the reward function are sampled from their posterior distribution, thus inducing exploration over the action space \cite{russo2018}. 
These sampled parameters are then used to form an estimated reward function and the action with the highest estimated reward is selected.

The hyper-parameters (e.g., the variance of the random effects) control the extent of pooling across users and across decision times. The right amount of pooling depends on the heterogeneity among users and the non-stationarity, which is often difficult to pre-specify.  Unlike other bandit algorithms in which the hyper-parameters are set at the beginning \cite{deshmukh2017multi,cesa2013gang,vaswani2017horde}, \ourapproach{} includes a procedure for updating the 
hyper-parameters online.  In particular, empirical Bayes \cite{carlin2010bayes} is used to update the hyper-parameters in the online setting, as more data becomes available.

\subsection{Problem formulation}
\label{sec: notation}
Consider an mHealth study which will recruit a total of $N$ users. 
\footnote{More generally, one can consider the setting where users become known to an algorithm over time. For example, users may open or delete accounts on an online shopping platform.} 
Let $i \in [N] = \{1, \dots, N\}$ be a user index.   
For each user, we use $k \in \{1, 2, \dots\}$ to index decision times, i.e., times at which a treatment could be provided.
Denote by $S_{i, k}$ the states/contexts at the $k^{th}$ decision time of user $i$.  
For simplicity, we focus on the case where the action is binary, i.e., $A_{{i, k}} \in \{0,1\}$.
The algorithm can be easily generalized to cases with more than two actions. After the action $A_{i, k}$ is chosen, the reward $R_{i, k}$ is observed.  Throughout the remainder of the paper, $S, A$ and $R$ are random variables and we use lower-case ($s$, $a$ and $r$) to refer to a realization of these random variables.

Below we consider a simpler setting where the parameters in the reward are assumed time-stationary. We discuss how to generalize the algorithm to the non-stationary setting in  \secref{sec:time}. The goal is to learn personalized treatment policies for each of the $N$ users. We treat this as $N$ contextual bandit problems 
as the reward function may differ between users. 
In mHealth settings this might occur due to the inability of sensors to record  users' entire contexts. 
\secref{sec:bandit_formulation} reviews two approaches for using  Thompson Sampling \cite{agrawal2012analysis} and \secref{sec:pooling_method} presents \ourapproach, our approach  for learning the treatment policy for any specific user. 

\subsection{Two Thompson Sampling instantiations}
\label{sec:bandit_formulation}

\begin{figure}
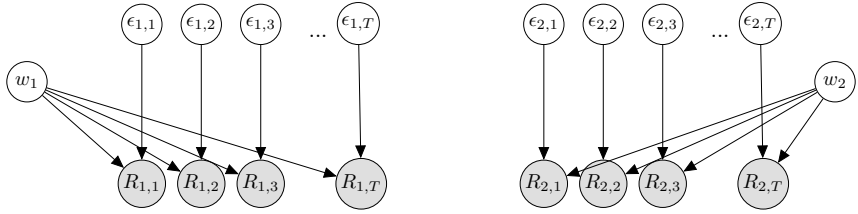

    \centering
    \resizebox{0.9\textwidth}{!}{%
       \tikz{
     \node[obs] (r_one) {$R_{1,1}$};%
     \node[obs,right of=r_one,xshift=1cm] (r_two) {$R_{1,2}$};%
     \node[obs,right of=r_two,xshift=1cm] (r_three) {$R_{1,3}$};%
     \node[right=of r_three] (dots_one) {...} ; %
     \node[obs,right of=dots_one] (r_t) {$R_{1,T}$};%
     \node[latent,above=of r_one,yshift=1.0cm] (noise_one) {${\epsilon_{1,1}}$}; %
     \node[latent,above=of r_two,yshift=1.0cm] (noise_two) {${\epsilon_{1,2}}$}; %
     \node[latent,above=of r_three,yshift=1.0cm] (noise_three) {${\epsilon_{1,3}}$}; %
     \node[above=of r_three,xshift=1.0cm,yshift=1.0cm] (dots) {...} ; %
     \node[latent,above=of r_three,xshift=1.7cm,yshift=1.0cm] (noise_T) {${\epsilon_{1,T}}$}; %
     \node[latent,above=of r_one,xshift=-2cm] (w_pop) {$w_1$}; %
     \node[obs,xshift=7cm] (r_one_2) {$R_{2,1}$};%
     \node[obs,right of=r_one_2,xshift=1cm] (r_two_2) {$R_{2,2}$};%
     \node[obs,right of=r_two_2,xshift=1cm] (r_three_2) {$R_{2,3}$};%
     \node[latent,above=of r_one_2,yshift=1.0cm] (noise_one_2) {${\epsilon_{2,1}}$}; %
     \node[latent,above=of r_two_2,yshift=1.0cm] (noise_two_2) {${\epsilon_{2,2}}$}; %
     \node[latent,above=of r_three_2,yshift=1.0cm] (noise_three_2) {${\epsilon_{2,3}}$}; %
      \node[above=of r_three_2,xshift=1.0cm,yshift=1.0cm] (dots_two) {...} ; %
     \node[right=of r_three_2] (dots_2) {...} ; %
     \node[obs,right of=dots_2] (r_t_2) {$R_{2,T}$};%
     \node[latent,above=of r_three_2,xshift=1.7cm,yshift=1.0cm] (noise_T_2) {${\epsilon_{2,T}}$}; %
     \node[latent,above=of r_three_2,xshift=3cm] (w_pop_2) {$w_2$}; %
     \edge {noise_one,w_pop} {r_one} ;
     \edge {noise_two,w_pop} {r_two} ;
     \edge {noise_three,w_pop} {r_three} ;
     \edge {noise_T,w_pop} {r_t} ;
     
      \edge {noise_one_2,w_pop_2} {r_one_2} ;
     \edge {noise_two_2,w_pop_2} {r_two_2} ;
     \edge {noise_three_2,w_pop_2} {r_three_2} ;
     \edge {noise_T_2,w_pop_2} {r_t_2} ;
     }
     }

     \caption{ Consider a setting with two users,  here we show the relationship between select random variables in our model: $R_{i,k}$ the reward for user $i$
     at decision time $k$, $\sigma^2_{\epsilon_{i, k}}$ the noise for user $i$ at time $k$ and $\w_{i}$ the latent weight vector for user $i$. 
     In \none{} we see that each user's parameters are independent.  Only the prior parameter values are shared, all else is updated independently. }
     \label{ps-plate}
     \end{figure}

First, consider learning the treatment policy separately per person. We refer to this approach as \none{}.
At each decision time $k$, we would like to select a treatment  $\activity_{i, k} \in \{0,1\}$ based on the context $\state_{i, k}$.  
We model the reward $\reward_{i, k}$  by a Bayesian linear regression model: for user $i$ and time $k$
\begin{eqnarray}
{
R_{i, k} = \phi(S_{i, k}, A_{i, k})^\transpose {\weightvector_{i}} + \epsilon_{i, k}},
\label{bayes_reg}
\end{eqnarray}
where $\context(s, a)$ is a pre-specified mapping from a context $s$ and treatment $a$ (e.g., those described in \secref{sec:implementation}), 
$\weightvector_i$ is a vector of weights which we will learn, and $\epsilon_{i, k} \sim \mathbf{N}(0, \sigma_{\epsilon}^2)$ is the error term.  
The weight vectors $\{\weightvector_i\}$ are assumed independent across users and to follow a common prior distribution $\weightvector_i \sim \mathbf{N}(\mu_\weightvector, \Sigma_\weightvector)$. See \figref{ps-plate} for a graphical representation of this approach.

Now at the $k^{th}$ decision time with the context $S_{i, k} = s$, \none{} selects the treatment $\activity_{i, k} = 1$ with probability  \begin{align}
\pi_{i, k} = \textrm{Pr} \{ \phi(s, 1) ^\transpose \tilde \weightvector_{i,k} > \phi(s, 0)^\transpose \tilde \weightvector_{i, k}\}  \label{prob}
\end{align}
where $\tilde \weightvector_{i, k}$ follows the posterior distribution of the parameters $\weightvector_i$ in the model (\ref{bayes_reg}) given the user's history up to the current decision time $k$.   We emphasize that in this formulation the posterior distribution of $\weightvector_i$ is formed based each user's own data.

\begin{figure}
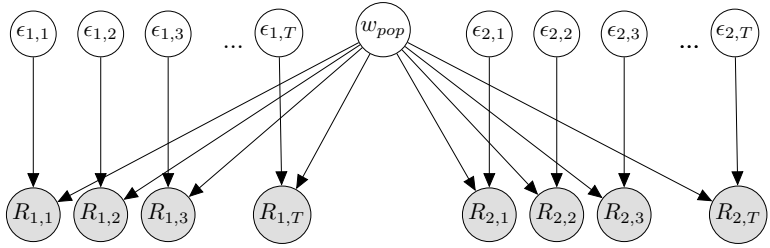

    \centering
    \resizebox{0.8\textwidth}{!}{%
       \tikz{
     \node[obs] (r_one) {$R_{1,1}$};%
     \node[obs,right of=r_one,xshift=1cm] (r_two) {$R_{1,2}$};%
     \node[obs,right of=r_two,xshift=1cm] (r_three) {$R_{1,3}$};%
     \node[right=of r_three] (dots_one) {...} ; %
     \node[obs,right of=dots_one] (r_t) {$R_{1,T}$};%
     \node[latent,above=of r_one,yshift=1.0cm] (noise_one) {${\epsilon_{1,1}}$}; %
     \node[latent,above=of r_two,yshift=1.0cm] (noise_two) {${\epsilon_{1,2}}$}; %
     \node[latent,above=of r_three,yshift=1.0cm] (noise_three) {${\epsilon_{1,3}}$}; %
     \node[above=of r_three,xshift=1.0cm,yshift=1.0cm] (dots) {...} ; %
     \node[latent,above=of r_three,xshift=1.7cm,yshift=1.0cm] (noise_T) {${\epsilon_{1,T}}$}; %
     \node[latent,above=of r_three,xshift=3.3cm,yshift=1.0cm] (w_pop) {$w_{pop}$}; %
     \node[obs,xshift=7cm] (r_one_2) {$R_{2,1}$};%
     \node[obs,right of=r_one_2,xshift=1cm] (r_two_2) {$R_{2,2}$};%
     \node[obs,right of=r_two_2,xshift=1cm] (r_three_2) {$R_{2,3}$};%
      \node[above=of r_three_2,xshift=1.0cm,yshift=1.0cm] (dots_two) {...} ; %
     \node[right=of r_three_2] (dots_2) {...} ; %
     \node[obs,right of=dots_2] (r_t_2) {$R_{2,T}$};
     \node[latent,above=of r_one_2,yshift=1.0cm] (noise_one_2) {${\epsilon_{2,1}}$}; %
     \node[latent,above=of r_two_2,yshift=1.0cm] (noise_two_2) {${\epsilon_{2,2}}$}; %
     \node[latent,above=of r_three_2,yshift=1.0cm] (noise_three_2) {${\epsilon_{2,3}}$}; %
      \node[above=of r_three_2,xshift=1.0cm,yshift=1.0cm] (dots_two) {...} ; %
     \node[latent,above=of r_three_2,xshift=1.7cm,yshift=1.0cm] (noise_T_2) {${\epsilon_{2,T}}$}; 

     \edge {noise_one,w_pop} {r_one} ;
     \edge {noise_two,w_pop} {r_two} ;
     \edge {noise_three,w_pop} {r_three} ;
     \edge {noise_T,w_pop} {r_t} ;
     \edge {noise_one_2,w_pop} {r_one_2} ;
     \edge {noise_two_2,w_pop} {r_two_2} ;
     \edge {noise_three_2,w_pop} {r_three_2} ;
     \edge {noise_T_2,w_pop} {r_t_2} ;

     }}

     \caption{
    Consider a setting with two users,  here we show the relationship between select random variables in our model: $R_{i,k}$ the reward for user $i$
     at decision time $k$, $\epsilon_{k}$ the noise at time $k$ and $w_{pop}$ the latent weight vector. 
     In \complete{} we see that each user's parameters are the same.  With each parameter update the weight vector for every user is also updated. }
     \label{complete-plate}
     \end{figure}

The opposite approach is to learn a common bandit model for all users. 
In this approach, the reward model is a single Bayesian regression model with no individual-level parameters: 
\begin{align}
\reward_{i, k}= \phi(\state_{i, k},\activity_{i, k})^\transpose \weightvector + \epsilon_{i, k}. \label{compete model}
\end{align}
where the common parameters, $\weightvector$, follows the prior distribution $\weightvector \sim \mathbf{N}(\mu_\weightvector, \Sigma_\weightvector)$. See \figref{complete-plate} for the graphical representation of this approach. We then use the posterior distribution of the weight vector $\weightvector$ to sample treatments for each user.  Here the posterior is calculated based on the available data from all users observed up to and including time $k$. 
This approach, which we refer to as \complete{}, may suffer from high bias when there is significant heterogeneity among users.

\subsection{Intelligent pooling across bandit problems}
\label{sec:pooling_method}
\ourapproach{} is an alternative to  the two approaches mentioned above.  Specifically,
in  \ourapproach{} data is pooled  across users in an adaptive way, i.e., when there is strong homogeneity  observed in the current  data, the algorithm will pool more from others than when there is strong heterogeneity.

\subsubsection*{Model specification}
\begin{figure}
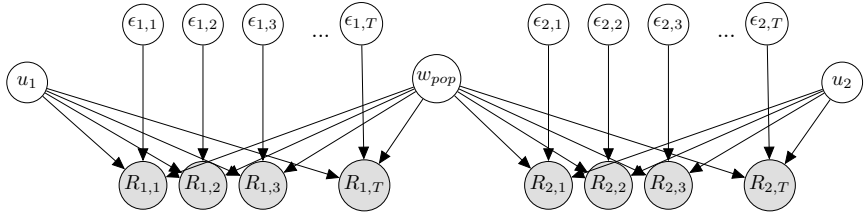

    \centering
    \resizebox{0.9\textwidth}{!}{%
       \tikz{
     \node[obs] (r_one) {$R_{1,1}$};%
     \node[obs,right of=r_one,xshift=1cm] (r_two) {$R_{1,2}$};%
     \node[obs,right of=r_two,xshift=1cm] (r_three) {$R_{1,3}$};%
     \node[right=of r_three] (dots_one) {...} ; %
     \node[obs,right of=dots_one] (r_t) {$R_{1,T}$};%
     \node[latent,above=of r_one,yshift=1.0cm] (noise_one) {${\epsilon_{1,1}}$}; %
     \node[latent,above=of r_two,yshift=1.0cm] (noise_two) {${\epsilon_{1,2}}$}; %
     \node[latent,above=of r_three,yshift=1.0cm] (noise_three) {${\epsilon_{1,3}}$}; %
     \node[above=of r_three,xshift=1.0cm,yshift=1.0cm] (dots) {...} ; %
     \node[latent,above=of r_three,xshift=1.7cm,yshift=1.0cm] (noise_T) {${\epsilon_{1,T}}$}; %
     \node[latent,above=of r_one,xshift=-2cm] (u_one) {$u_1$}; %
     \node[latent,above=of r_three,xshift=3.0cm] (w_pop) {$w_{pop}$}; %
     \node[obs,xshift=7cm] (r_one_2) {$R_{2,1}$};%
     \node[obs,right of=r_one_2,xshift=1cm] (r_two_2) {$R_{2,2}$};%
     \node[obs,right of=r_two_2,xshift=1cm] (r_three_2) {$R_{2,3}$};%
     \node[latent,above=of r_one_2,yshift=1.0cm] (noise_one_2) {${\epsilon_{2,1}}$}; %
     \node[latent,above=of r_two_2,yshift=1.0cm] (noise_two_2) {${\epsilon_{2,2}}$}; %
     \node[latent,above=of r_three_2,yshift=1.0cm] (noise_three_2) {${\epsilon_{2,3}}$}; %
      \node[above=of r_three_2,xshift=1.0cm,yshift=1.0cm] (dots_two) {...} ; %
     \node[right=of r_three_2] (dots_2) {...} ; %
     \node[obs,right of=dots_2] (r_t_2) {$R_{2,T}$};%
     \node[latent,above=of r_three_2,,xshift=1.7cm,yshift=1.0cm] (noise_T_2) {${\epsilon_{2,T}}$}; %
     \node[latent,above=of r_three_2,xshift=3cm] (u_two) {$u_2$}; %
   
     \edge {noise_one,u_one,w_pop} {r_one} ;
     \edge {noise_two,u_one,w_pop} {r_two} ;
     \edge {noise_three,u_one,w_pop} {r_three} ;
     \edge {noise_T,u_one,w_pop} {r_t} ;
     
      \edge {noise_one_2,u_two,w_pop} {r_one_2} ;
     \edge {noise_two_2,u_two,w_pop} {r_two_2} ;
     \edge {noise_three_2,u_two,w_pop} {r_three_2} ;
     \edge {noise_T_2,u_two,w_pop} {r_t_2} ;

     }}

     \caption{ Consider a setting with two users, here we show the relationship between select random variables in our model: $R_{i,k}$ the reward for user $i$
     at decision time $k$, ${\epsilon_{i,k}}$ the noise for user $i$ at time $k$, $w_{pop}$ the latent weight vector and $u_i$ the random effect for user $i$. In \ourapproach{} we see that some parameters ($w_{pop}$) are shared across the population which others ($u_i$) are user specific.\label{ip-plate}}
     \end{figure}

We model the reward associated with taking action $A_{i, k}$ for user $i$ at decision time $k$ by the linear model (\ref{bayes_reg}). Unlike \none{} where the person-specific weight vectors $\{w_i, i \in [N]\}$ are assumed to be independent to each other, \ourapproach{} imposes structure on the  $\weightvector_{i}$'s, in particular, a random-effects  structure \cite{raudenbush2002hierarchical,laird1982random}:
\begin{eqnarray}
{
\weightvector_{i} = \weightvector_{pop} + u_i 
,\label{random_effect} 
}
\label{randomeffect1}
\end{eqnarray}
where $\weightvector_{pop}$ is a population-level parameter and $u_i$ is a  \textit{random effect} that 
represents the person-specific deviation from $\weightvector_{pop}$ for user $i$. 
The extent to which the posterior means for $\weightvector_{pop}$ and $ u_i$ are based on user $i$'s data
relative to the population depends on the variances of the random effects (for a stylized example of this see \secref{subsec:intuition}). 
In \secref{sec:time} we show how we can modify this structure to include time-specific parameters, or a time-specific random effect.  A graphical representation for \ourapproach{} is shown in \figref{ip-plate}.

We assume the prior on 
 $\weightvector_{pop}$ is Gaussian with prior mean $\mu_{\weightvector}$ and variance $\Sigma_\weightvector$.  $u_i$ is also assumed to be Gaussian with mean $\mathbf{0}$ and covariance $\Sigma_u$. Furthermore, we assume $u_i \independent u_j$ for $i \neq j$ and $\weightvector_{pop} \independent \{u_i\} $ .
The prior parameters $\mu_{\weightvector}, \Sigma_\weightvector$
 as well as the  variance of the random effect $\Sigma_u$, and the residual variance $\sigma_\epsilon^2$ are hyper-parameters. In (\ref{randomeffect1}), there is a  the random effect, $u_i$ on each element of $\weightvector_i$.
 In practice, one can use domain knowledge to specify which of the parameters should include random effects; this will be the case in the feasibility study described in Section 6.
Conditioned on the latent variables $(\weightvector_{pop}, u_i)$, as well as the current context and action, 
the expected reward is $$E[R_{i,k}|w_{pop}, u_i, S_{i,k}=s, A_{i,k}=a]= \phi(s,a)^T (w_{pop}+ u_i).$$

\subsubsection*{Model connections to Gaussian Processes}
Under the Gaussian assumption on the distribution of the reward and prior, the Bayesian linear model of the reward (\ref{bayes_reg}) together with the random effect model (\ref{randomeffect1}) can be viewed as an example of Gaussian Process with a special kernel (see \eqnref{our_kernel}).   We use this connection to derive the posterior distribution and facilitate the hyper-parameter selection. An additional advantage of viewing the Bayesian mixed effects model as a Gaussian Process model
is that we can now flexibly redesign our reward model simply by introducing 
new kernel functions. Here, we assume linear model with a person-specific random effects. In \secref{sec:time}
we discuss a generalization to time-specific random effects. Additionally, one could adopt non-linear kernels and incorporate more complex structures on the reward function.

\subsubsection*{Posterior distribution of the weights on the feature vector}
In the setting where both the prior and the linear model for the reward follow a Gaussian distribution, the posterior distribution of $\weightvector_{i}$ follows a Gaussian distribution and there are analytic expressions for these updates, as shown  in \cite{williams2006gaussian}. Below we provide the explicit formula of the posterior distribution based on the connection to a Gaussian Process regression.  Suppose at the time of updating the posterior distribution, the available data collected from all current users is $\D$, 
where $\D$ consists of $n$ tuples of state, action, reward and user index $x = (s, a, r, i)$. 
The
mixed effects 
model (Eqns.~\ref{bayes_reg} and \ref{randomeffect1})  induces a kernel function $K$. 
For any two tuples in $\D$, e.g., $x_l = (s_{l}, a_{l}, r_{l}, i_l), l = 1, 2$
\begin{align}
K_{}(x_1, x_2) & =\phi(s_1, a_1)^\transpose  (\Sigma_\weightvector + \indicator{i_1 = i_2} \Sigma_{u} )  \phi(s_2, a_2).
\label{our_kernel}
\end{align}

Note that the above kernel depends on $\Sigma_\weightvector$ and $\Sigma_u$ (one of the hyper-parameters that will be updated using empirical Bayes approach; see below).  The kernel matrix $\mathbf{K}$ is of size $n \times n$ and each element is the kernel value between two tuples in $\D$. The posterior mean and variance of $\weightvector_{i}$ given the currently available data $\D$ can be calculated by
\begin{align}
\begin{split}
\hat{\weightvector}_{i}  &= \mu_\weightvector + M_{i}^\transpose  (\mathbf{K} + \sigma_{\epsilon}^2 I_{n})^{-1} \tilde R_{n}\\
\Sigma_{i}  &= \Sigma_\weightvector + \Sigma_u  - M_{i}^\transpose  (\mathbf{K} + \sigma_{\epsilon}^2 I_{n})^{-1} M_{i}
\end{split}
\label{post cal}
\end{align}
where 
$\tilde R_{n}$ is the vector of the rewards centered by the prior means, i.e., each element corresponds to a tuple $(s, a, r, j)$ in $\D$ given by $r - \phi(s, a)^\transpose \mu_{\weightvector}$, and 
 $M_{i}$ is a matrix of size $n$ by $p$ (recall $p$ is the length of $\weightvector_i$), with each row corresponding to a tuple $(s, a, r, j)$ in $\D$ given by $\phi(s, a)^\transpose (\Sigma_\weightvector+ \indicator{j = i} \Sigma_{u} )$.

\subsubsection*{Treatment selection}
To select a treatment for user $i$ at the $k^{th}$ decision time, we use the posterior distribution of $\weightvector_{i}$  formed at the most recent update time $T$.   That is, for the context $S_{i, k}$ of user $i$ at the $k^{th}$ decision time,  \ourapproach{} selects the treatment $A_{i,k} = 1$  with the probability calculated in the same formula as in (\ref{prob}) but with a different posterior distribution as discussed above.

\subsubsection*{Setting hyper-parameter values}
Recall that the algorithm requires the hyper-parameters $\mu_{\weightvector}, \Sigma_\weightvector$, $\Sigma_u$, and $\sigma_\epsilon^2$. 
The prior mean $\mu_{\weightvector}$  and variance $\Sigma_\weightvector$ of the population parameter $\weightvector_{pop}$ can be set according to previous data or domain knowledge (see Section \ref{sec:clinical} for a discussion on how the prior distribution is set in the feasibility study).  
As we mention in Section \ref{sec: overview}, the variance components in the mixed effects model impact how the users pool the data from others (see \secref{subsec:intuition} for a discussion) and might be difficult to pre-specify.  \ourapproach{} uses, at the update times, the empirical Bayes \cite{carlin2010bayes} approach to choose/update $\lambda = (\Sigma_u, \sigma_\epsilon^2)$ based on the currently available data. To be more specific, suppose at the time of updating the hyper-parameters, the available data  is  $\D$.  We choose $\lambda$ to 
 maximize $l(\lambda | \D)$, the marginal log-likelihood of the observed reward, marginalized over the population parameters $\weightvector_{pop}$ and the random effects $u_i$. The marginal log-likelihood $ l(\lambda | \D)$ can be expressed as
\begin{align}
\label{marginal likelihood}
\begin{split}
 l(\lambda | \D)  = -\frac{1}{2} \Big\{ \tilde R_{n}^\transpose [\mathbf{K}(\lambda) & + \sigma_\epsilon^2 I_{n}]^{-1} \tilde R_{n} 
   + \log \det [\mathbf{K}(\lambda)  + \sigma_\epsilon^2 I_{n}] + n \log(2\pi) \Big\}
\end{split}
\end{align}
where  $\mathbf{K}(\lambda)$ is the kernel matrix as a function of parameters $\lambda = (\Sigma_{u}, \sigma_{\epsilon}^2)$. 
The above optimization can be efficiently solved using existing Gaussian Process regression packages; see \secref{sec:implementation} for more details.

\begin{algorithm}
\caption{ \ourapproach{}\label{pooledalg}} 
\begin{algorithmic}[1]
 \STATE Let $\mathcal{T}$
be a set of all times at which the algorithm
might deliver a treatment or perform a parameter update.
\STATE \begin{ppl}Set $\hat{\weightvector}_{i, 0} = \mu_\weightvector,  \Sigma_{i,0} = \Sigma_\weightvector+\Sigma_u$  for all $i$ and $\mathcal{D}=\{\}$. \end{ppl}  
 \FOR {all $t$  $\in  \mathcal{T}$}
    \IF{$t$ is a decision time}
	\STATE \begin{ppl}Receive user index $i$ and  decision time index $k$\end{ppl}
    \STATE\begin{ppl} Collect  state variable $\state_{i,k}$\end{ppl}
    \STATE \begin{ppl}Calculate randomization probability \end{ppl}$\pi_{i, k}= \textrm{Pr}_{\tilde\weightvector \sim   \mathbf{N}(\hat{\weightvector}_{i} , \Sigma_{i})} \{ \phi(S_{i, k}, 1) ^\transpose \tilde \weightvector > \phi(S_{i, k}, 0)^\transpose \tilde\weightvector\}$
    \STATE \begin{ppl}Sample  treatment $\activity_{i,k} \sim  \operatorname{Bern} \left({\pi_{i, k}}\right)$\end{ppl} 
    \STATE \begin{ppl} Collect  reward $\reward_{i, k}$ \end{ppl} 
    \STATE $\D \leftarrow \D \cup \{S_{i,k}, A_{i,k}, R_{i,k}, i\}$
     \ENDIF 
    \IF{$t$ is an update time}
    \STATE \begin{ppl}Update the hyper-parameters: $\hat \lambda = \argmax l(\lambda | \D)$ in Eqn \ref{marginal likelihood}\end{ppl} 
   \STATE \begin{ppl}Update the posterior mean and covariance $\hat\weightvector_{i},\Sigma_{i}$ for all $i$ in $\D$ by Eqns \ref{post cal} with $\hat \lambda$ \end{ppl} 
  \ENDIF 
 \ENDFOR
\end{algorithmic}

\end{algorithm}

\subsection{Intuition for the use of random effects}
\label{subsec:intuition}

\ourapproach{} uses  random effects to adaptively pool users' data based on the degree to which  users exhibit heterogeneous rewards. That is, the person-specific random effect should outweigh the population term if users are highly heterogeneous. If users are highly homogeneous,  the person-specific random effect should be outweighed by the population term.  The amount of pooling is controlled by the hyper-parameters, e.g., the variance components of the random effects. 

To gain intuition, we consider a simple setting where the feature vector $\phi$ in the reward model (Eqn. \ref{bayes_reg}) is one-dimensional (i.e., $p =1$) and there are only two users (i.e., $i=1,2$).  Denote the prior distributions of population parameter $\weightvector_{pop}$ by $\mathbf{N}(0, \sigma_{\weightvector}^2)$ and the random effect $u_i$ by $\mathbf{N}(0,  \sigma_u^2)$.  Below we investigate how the hyper-parameter (e.g., $\sigma_u^2$ in this simple case) impacts the posterior distribution.  

	\begin{figure}
		\centering
		\includegraphics[width=.7\linewidth]{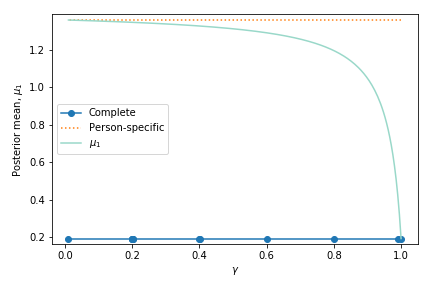}

		\caption{The posterior mean of $\weightvector_i$, $\hat\weightvector_1$. As the variance of random effect $\sigma_u^2$ decreases, $\gamma$ increases and the posterior mean approaches the population-informed estimation (\complete) and departs from the person-specific estimation (\none).  		\label{figexample} }
	\end{figure}

Let $k_i$ be the number of decision time of user $i$ at an updating time. 
In this simple setting, the posterior mean of $\hat\weightvector_1$ can be calculated explicitly: 
	\[
\hat\weightvector_{1} =   
	 \frac{[\delta \gamma + (1-\gamma^2) C_2] Y_1 + \delta \gamma^2 Y_2}{(1-\gamma^2)C_1 C_2  + \delta \gamma(C_1 + C_2) + (\delta \gamma)^2}
	\]
where  for $i=1,2$, $C_i = \sum_{k=1}^{k_i} \phi(A_{i, k}, S_{i, k})^2$,  $Y_i = \sum_{k=1}^{k_i}  \phi(A_{i, k}, S_{i, k}) R_{i, k}$, $\gamma =  \sigma_\weightvector^2/(\sigma_\weightvector^2 + \sigma_u^2)$ and $\delta = \sigma_\epsilon^2/\sigma_\weightvector^2$.	Similarly, the posterior mean of $\weightvector_2$ is given by
\[
\hat\weightvector_{2} =
\frac{[\delta \gamma + (1-\gamma^2) C_1] Y_2 + \delta \gamma^2 Y_1}{(1-\gamma^2)C_1 C_2  + \delta \gamma(C_1 + C_2) + (\delta \gamma)^2}
\]

 When $\sigma_u^2 \goes 0$ (i.e., the variance of random  effect goes to 0), we have $\gamma \goes 1$ and both posterior means ($\hat\weightvector_{1}, \hat\weightvector_{2}$) approach the posterior mean under \complete ~(Eqn \ref{compete model}) using prior $\mathbf{N}(0,  \sigma_\weightvector^2)$

 $$\hat\weightvector_{1}, \hat\weightvector_{2} \goes \frac{Y_1 + Y_2}{C_1 + C_2 + \delta}.$$

  Alternatively, when $\sigma_u^2 \goes \infty$, we have $\gamma \goes 0$ and the posterior means ($\hat\weightvector_{1}, \hat\weightvector_{2}$) each approach their respective posterior means under \none{} (Eqn \ref{bayes_reg}) using a non-informative prior
	$$
	\hat\weightvector_1 \goes \frac{Y_1}{C_1}, ~\hat\weightvector_2 \goes \frac{Y_2}{C_2}.
	$$
 \figref{figexample} illustrates that when $\gamma$ goes from 0 to $1$, the posterior mean $\hat\weightvector_i$ smoothly transitions from the population estimates to the person-specific estimates.

 \subsection{Regret}
 We prove a regret bound for a modification of \ourapproach{} similar to that in  \cite{agrawal2012analysis,vaswani2017horde} in a simplified setting.  Further details are provided in Appendix \ref{sec:proof_details}.  Let $d$ be the length of the weight vector $w_i$ in the Bayesian mixed effects model of the reward in \eqnref{bayes_reg}. 
Recall that $\Sigma_w$ is the prior covariance of the weight vector $\weightvector_{pop}$, $\Sigma_u$ is the covariance of the random effect $u_i$ and $\sigma_\epsilon^2$ is the variance of the error term. 
Let $K_i$ be the number of decision times for user $i$ up to a given calendar time and $T = \sum_{i=1}^N K_i$ be the total number of decision times encountered by all $N$ users in the study up to the calendar time.  We define the regret of the algorithm after $T$ decision times by 
$\mathcal{R}(T) = \sum_{i=1}^N \sum_{k=1}^{K_i} \max_{a} \phi(S_{i,k}, a)^T w_{i} - \phi(S_{i,k}, A_{i, k})^T w_{i}$.

\begin{theorem}
\label{regret_bound}
With probability $1-\delta$, where $\delta \in (0,1)$ the total regret of the modified Thompson Sampling with \ourapproach{} after $T$ total number of decision times is: \\
$$\mathcal{R}(T) = \mathcal{\tilde{O}}\Bigg(dN\sqrt{T}\sqrt{\log\Big(\frac{( \trace{\Sigma_\weightvector}+\trace{\Sigma_u}+\trace{ \Sigma_u^{-1})}}{d}+\frac{T}{\sigma_\epsilon^2dN} \Big) \log{\frac{1}{\delta}}}\Bigg)$$
\end{theorem}

\paragraph{Remark}
Observe that, up to logarithmic terms, this regret bound is $\tilde O(dN\sqrt T)$. 
Recall that \cite{vaswani2017horde} introduces a similar regret bound for a Thompson Sampling 
algorithm which utilizes user-similarity information. 
 The bound from \cite{vaswani2017horde}, $\tilde O(dN\sqrt{T/\lambda})$, 
 additionally depends on a hyper-parameter 
 $\lambda$ that is not included in our model. In \cite{vaswani2017horde}, $\lambda$ controls the strength 
of prior user-similarity information. 
Instead of introducing a hyper-parameter our
model follows a mixed effects Bayesian structure
which allows user similarities (as expressed in the extent to which users' data is pooled)
to be updated with new data.
Thus, in certain regimes of hyper-parameter $\lambda$, 
\ourapproach{} will incur much smaller regret, as demonstrated
empirically in \secref{sec:empirical_eval}.

\section{Experiments}
\label{sec:experimental_design}
This work was conducted to prepare for deployment of  \ourapproach{} in a live trial.
Thus, to evaluate \ourapproach{} we construct a simulation environment from a precursor trial,  \HSVone{}\cite{klasnja2015microrandomized}. This simulation allows us to evaluate the proposed algorithm under various   settings that may arise in  implementation. 
 For example, heterogeneity in the observed rewards may be due to unknown subgroups across which users' reward functions differ. Alternatively,  this heterogeneity may vary across users in a more continuous manner.  We consider both scenarios in simulated trials. 
 In Sections \ref{sec:sim}-\ref{sec:empirical_eval} we evaluate the performance of \ourapproach{} against baselines and a state-of-the-art algorithm. 
In \secref{sec:clinical} we assess feasibility of \ourapproach{} in a pilot deployment  in a clinical trial.

 \subsection{Simulation environment}
 \label{sec:sim}
\HSVone{} was a 6-week micro-randomized trial of an Android-based physical activity intervention with 41 sedentary adults. The intervention consisted of two \textit{push} interventions: planning and contextually-tailored activity suggestions. Activity suggestions acted as action cues and were designed to provide users with actionable options for engaging in short bouts of activity in their current situation. The content of the suggestions was tailored based on the users' location, weather, time of day, and day of the week. For each individual, on each day of the study, the HeartSteps system randomized whether or not to send an activity suggestion five times a day. The intended outcome of the suggestions---the proximal outcome used to evaluate their efficacy---was the step count in the 30 minutes following suggestion randomization. 

\HSVone{} data was used to construct all features within the environment, and to guide choices such as how often to update the feature values.
 Recall that $S_{i,k}$ and $R_{i,k}$ denote the context features and reward of user $i$ at the $k^{th}$ decision time. The reward is the log step counts in the thirty minutes immediately following a decision time.
In \HSVone{} three treatment actions were considered:
 $A_{i,k}=1$ corresponded to a smartphone notification containing an activity suggestion designed to take 3 minutes to perform, $A_{i,k}=0$ corresponded to a smartphone notification containing an anti-sedentary message  designed to take approximately 30 seconds to perform and $A_{i,k}=-1$ corresponded to not sending a message. However, in the simulation only the actions $1,0$ are considered.
\begin{figure}
\centering

\begin{minipage}[b]{.95\columnwidth}
    \centering\includegraphics[width=12.0cm]{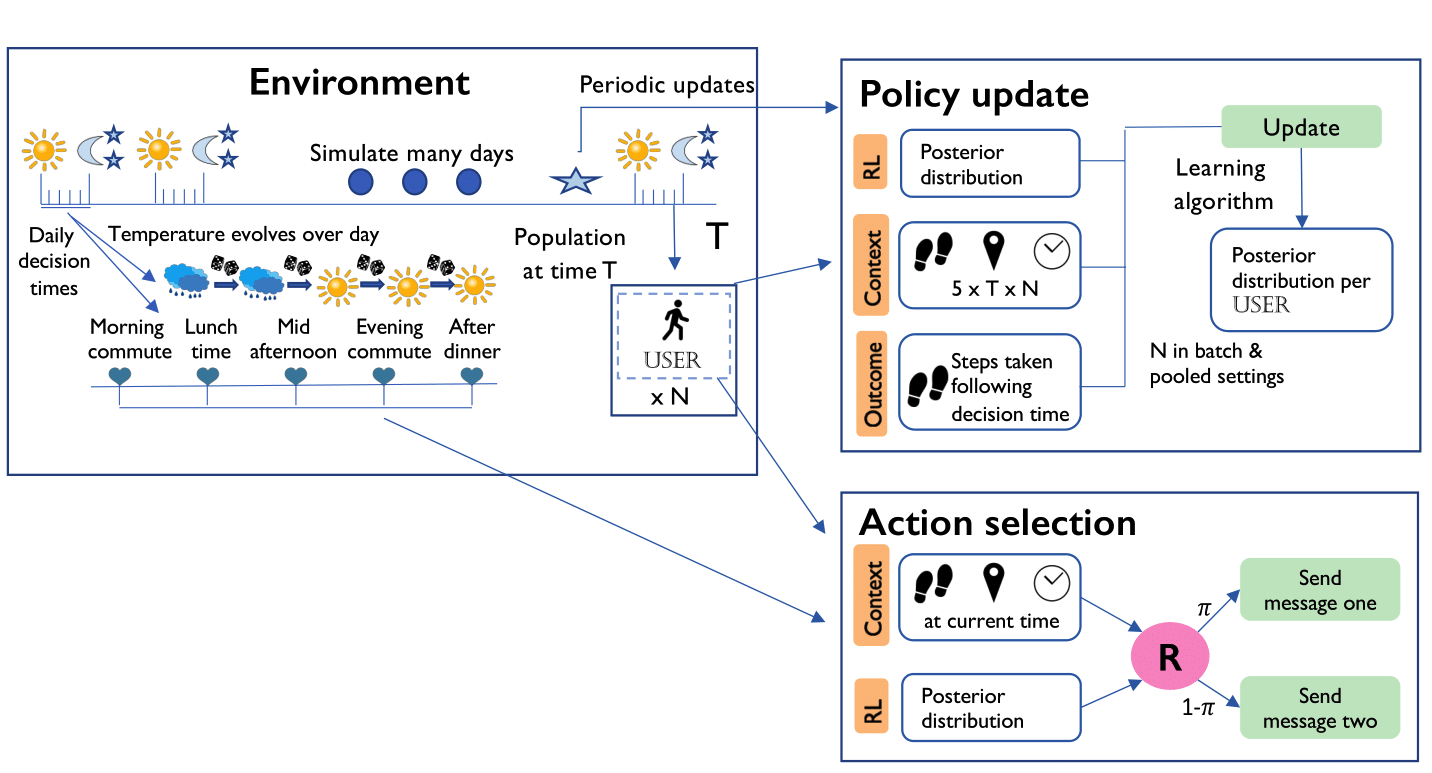}
    \caption{ Contextual features for a simulated \patient{} are composed of both general environmental features (such as time of day) and individual features (such as location). At
    decision times  a simulated user receives a message determined by the current treatment policy. Periodically this policy is updated according to a learning algorithm which outputs a new posterior distribution for each \patient. \label{sim_pic}}
    \end{minipage}
\end{figure}
\figref{sim_pic} describes the simulation while \tabref{user_states} describes  context features and rewards. 
Each context feature in \tabref{user_states} was constructed from  \HSVone{} data.  
For example, we found that in \HSVone{} data
splitting participants' prior 30 minute step count  into the two categories of high or low best explained the reward. Additional details about this process are included in \secref{sup:features}.

The temperature and location are updated throughout a simulated day according to probabilistic transition functions constructed from  \HSVone{}.
The step counts for a simulated user are generated from  participants in \HSVone{} as follows.   
We construct a one-hot feature vector  containing the group-ID of a participant, the time of day, the day of the week, the temperature, the preceding activity level, and the location.
 Then for each possible realization of  the one-hot encoding we calculate the empirical mean  and empirical standard deviation of all step counts observed in  \HSVone{}.
 The corresponding  empirical mean  and empirical standard deviation from \HSVone{} form
 $\mu_{\state_{i,k}}$  $\sigma_{\state_{i,k}}$ respectively.  
At each 30 minute window, if a treatment is not delivered 
step counts are generated according to 
\begin{equation}R_{i,k} = \mathbf{N}(\mu_{\state_{i,k}},\sigma^2_{\state_{i,k}}).\label{not_avail}\end{equation}

\begin{table}
\small
\centering
\resizebox{1.0\columnwidth}{!}{%
\begin{tabular}{|p{4.0cm}p{4.0cm}p{2.5cm}|}
\hline
    \multicolumn{3}{|c|}{ 
\normalsize{ \textbf{State ($S$) Features}}} \\
\hline
    \multicolumn{1}{|l|}{\textbf{Name}}  &  \multicolumn{1}{l|}{\textbf{Value} }&  \multicolumn{1}{l|}{\textbf{ \makecell[l]{ \textbf{\patient}\\ Specific}  }}\\
\hline
Time of day & \makecell[l]{Morning  9:00 and 15:00 (0) \\ Afternoon   15:00 and 21:00 (1)} & \hspace{3mm}No   \\
\hline
Day of the week & Weekday (0) or Weekend (1)  & \hspace{3mm}No \\
\hline
Temperature& Cold (0) or Hot (1) & \hspace{3mm}No \\
\hline
  Preceding activity level& Low (0) or High (1)& \hspace{3mm}Yes \\
Location & Other (0) or Home/work (1)& \hspace{3mm}Yes \\
Intercept & 1 & \hspace{3mm}Yes \\
\hline
    \multicolumn{3}{|c|}{
\normalsize{\textbf{Reward}} } \\
    \hline
Step count & Continuous on log scale& \hspace{3mm}Yes \\
\hline
\end{tabular}
}
\caption{\textmd{The value used in encoding each feature is shown in parentheses. For example cold (0) indicates that cold is coded as a 0 wherever this feature is used. 
A user's state is described as $S_{i, k} = \{1, \text{time of day}, \text{day of the week}, \text{preceding activity level},  \text{location}\} $}. 
}
\label{user_states}
\end{table}

\textbf{\populationgen{}} 
This model, which we denote \populationgen{}, allows us to compare the performance of the approaches under different levels of population heterogeneity. 
The step count after a decision time is a modification of \eqnref{not_avail} to reflect the interaction between  context and treatment on the reward and heterogeneity in treatment effect.
Let $\beta$ be a vector of coefficients of $S_{i,k}$ which weigh the relative contributions of the entries of $S_{i,k}$ that interact with treatment on the reward. 
The magnitude of the entries of $\beta$  are set using \HSVone{}. Step counts ($R_{i,k}$) are generated  as 
  \begin{equation}R_{i,k} = \mathbf{N}(\mu_{\state_{i,k}},\sigma^2_{\state_{i,k}})+ A_{i,k}(\state_{i,k}^T\beta_{i} + Z_i). \label{avail} 
  \end{equation}

The inclusion of $Z_i$ will allow us to evaluate the relative performance of each approach under different levels of population heterogeneity. 
Let $\beta^l_i$ be the entry in $\beta_i$ corresponding to  the location term for the $i^{th}$ user. 
We consider three scenarios (shown in \tabref{table:Z}) to generate $Z_i$, the person-specific effect, and $\beta^l_i$ the location-dependent  effect. 
The performance of each algorithm under each scenario will be analyzed in \secref{sec:empirical_eval}.  In the smooth scenario, $\sigma$ is equal to the standard deviation of the observed treatment effects $[f(S_{i,k})^\transpose \beta\ :\ S_{i,k} \in \HSVone{}]$. The settings for all $Z_i$ and $\beta^l_i$ terms are discussed in \secref{sup:features}.

In the bi-modal scenario each simulated user is  assigned a base-activity level: low-activity users (group 1) or high-activity users (group 2). 
When a simulated user joins the trial they are placed into either group one or two with equal probability. 
Whether or not it is optimal to send a treatment (an activity suggestion) for user $i$ at their $k^{th}$ decision time depends both on their context, and on the values of  $z_1,\beta^l_1$ and $z_2,\beta^l_2$.
The values of  $z_1,\beta^l_1$ and $z_2,\beta^l_2$ are set so that for all users in group 1, it is optimal to send a treatment under 75\% of the contexts they will experience. Yet for all users in group 2, it is only optimal to send a treatment under 25\% of the contexts they will experience. Group membership is not known to any of the algorithms. 
The settings for all values in \tabref{table:Z} are included in \secref{sup:features}.\\

\begin{table}
\centering	
\resizebox{\columnwidth}{!}{%
\begin{tabular}{|p{1.5cm |}|p{1.5cm |}|p{1.5cm |}||}
\hline
   \multicolumn{1}{|c|}{Homogeneous}&\multicolumn{1}{c|} {Bi-modal } & \multicolumn{1}{c|}{Smooth} \\
\hline
    \multicolumn{1}{|c|}{$Z^i =0$ $\beta^l_i$=0} &   \multicolumn{1}{r|}{ $ Z_i,\beta^l_i = \begin{cases}
      z_1, \beta^l_1 & \text{if}\ i \in \text{group one} \\
      z_2, \beta^l_2&  \text{if}\ i \in  \text{group two}
    \end{cases}$ } &  \multicolumn{1}{r|}{ $Z_i \sim \mathcal{N}(0,\sigma^2)$ $\beta^l_i\sim \mathcal{N}(0,\sigma_l^2)$} \\
    \hline
\end{tabular}}
\caption{\textmd{Settings for Z in three cases of homogeneous, bimodal and smoothly varying populations. } \label{table:Z}}
\end{table}

 \subsection{Model for the reward function in \ourapproach{}}
 \label{sec:implementation}
In \secref{sec:approach} we introduced the feature vector  $\phi(\state_{i,k},\activity_{i,k}) \in \R^p$. 
This vector is used in  the model for the reward and
transforms a user's contextual state variables $\state_{i,k}$ 
and the action $\activity_{i,k}$ as follows:

\begin{equation} \label{phi_feature_vector}
\begin{split}
\phi(\state_{i,k},A_{i,k})^T  = & \big( \state_{i,k}^T, \pi_{i,k} \state_{i,k}^T, (A_{i,k}-\pi_{i,k})\state_{i,k} \big),
\end{split}
\end{equation}
where $S_{i, k} = \{1, \text{time of day}, \text{day of the week}, \text{preceding activity level},  \text{location}\}$.
Recall that the bandit algorithms produce $\pi_{i,k}$ which is the probability that $A_{i,k}=1$. 
The inclusion of the term $\small{(A_{i,k}-\pi_{i,k})\state_{i,k}}$ is motivated by \cite{liao2016sample,boruvka2018assessing,greenewald2017action}, who demonstrated that 
action-centering can protect against mis-specification in the baseline effect (e.g., the expected reward under the action 0).   
 In $\HSVone$ we observed that  users varied in their overall responsivity and that a user's location was related to their responsivity. In the simulation, we assume the person-specific random effect  on four parameters in the reward model (i.e., the coefficients of terms in $\state$ involving the intercept and location).

 Finally, we constrain the randomization probability to be within [0.1, 0.8] to ensure continual learning. The update time for the hyper-parameters is set to be every 7 days.
All approaches are implemented in Python and we implement GP regression with the software package GPytorch \cite{gardner2018gpytorch}.

  \subsection{Simulation results}
 \label{sec:empirical_eval}
In this section, we compare the use of mixed effects model for the reward function in $\ourapproach$ to  two standard methods used in mHealth, $\complete$ and $\none$ from  \secref{sec:bandit_formulation}.  Recall that  \ourapproach{} includes person-specific random effects, as described in \eqnref{random_effect}. 
In $\none$,  all users are assumed to be different and there is no pooling of data and in 
 $\complete$, we treat all users the same and learn one set of parameters across the entire population.

Additionally, to assess \ourapproach's ability to pool across users we compare our approach to Gang of Bandits  \cite{cesa2013gang}, which we refer to as \hob. 
As this model requires a relational graph between users, we  construct a graph using the generative model (\ref{avail}) and Table~\ref{table:Z} connecting users according to each of the three settings: homogeneous, bi-modal and smooth.
For example, with knowledge of the generative model users can be connected to other users as a function of their $Z_i$ terms. 
As we will not have true access to the underlying generative model in a real-life setting we distort the true graph to reflect this incomplete knowledge.
That is we add ties to dissimilar users at 50\% of the strength of the ties between similar users.

From the generative model (\ref{avail}), the optimal action for user $i$ at the $k^{th}$ decision time 
is $a^*_{i,k} = \indicator{S_{i,k}^T\beta_{i}^ *+Z_i \geq 0}$.  The  regret is
\begin{equation}
\text{regret}_{i,k}=|S_{i,k}^T\beta_{i}^ *+Z_i|  \indicator{a^*_{i,k}\neq A_{i,k}}  \label{eqn:regret}
\end{equation}
where $\beta^*_i$ is the optimal $\beta$ for the $i^{th}$ user.

In these simulations each trial has  32 users. 
Each user remains in the trial for 10 weeks and the entire length of the trial is 15 weeks, where the last cohort joins in week six.  The number of users who join each week is a function of the recruitment rate observed in $\HSVone$. In all settings we run 50 simulated trials. 

\begin{figure}
\centering
\begin{minipage}[b]{0.95\linewidth}
    \centering\includegraphics[width=8.0cm]{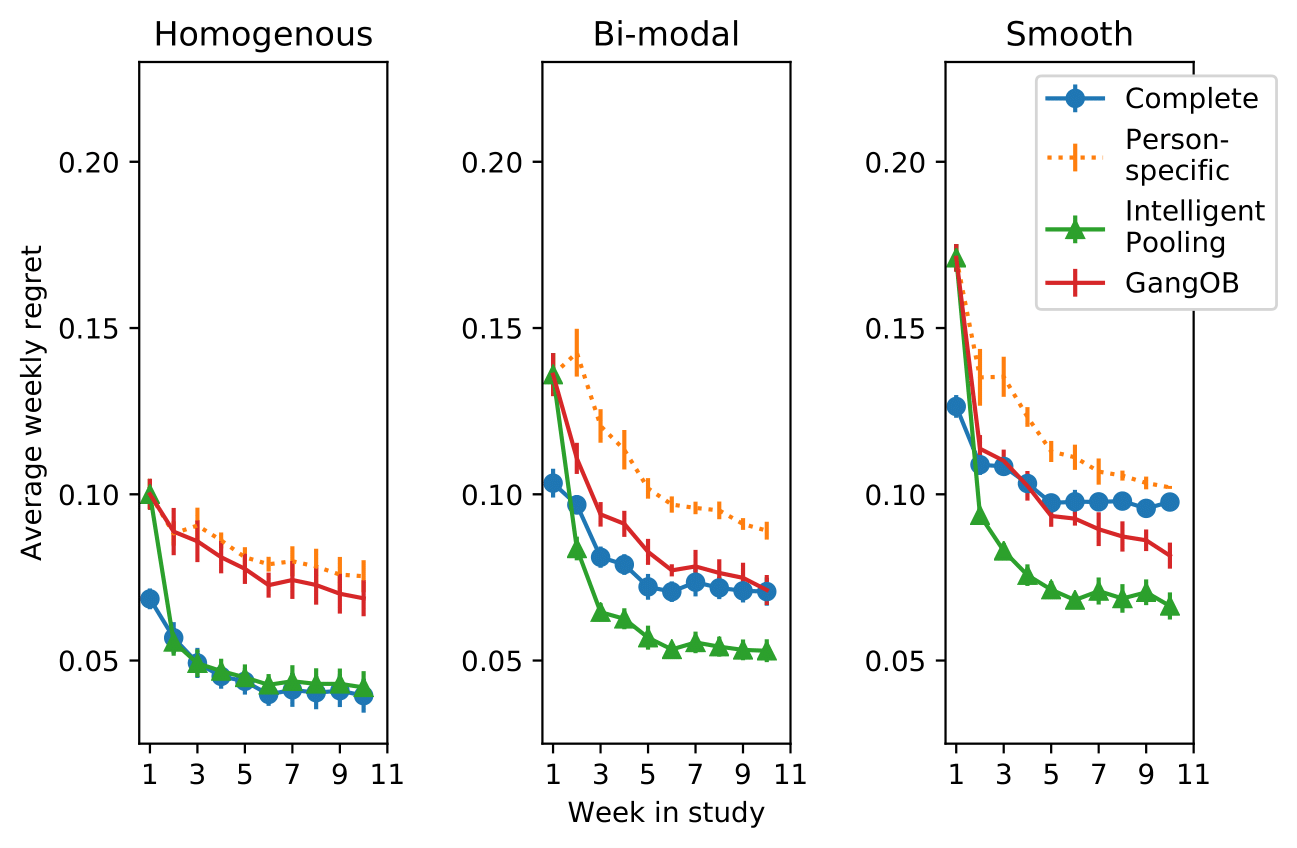}
    \caption{
\textbf{\populationgen{} generative model} 
 Regret averaged across all users for each week in the trial, i.e. average regret of all users in their first week of the trial.}\label{regret}
    \end{minipage}
  \end{figure}

First,  \figref{regret} provides the regret  averaged across all users across 50 simulated trials where the reward distribution follows (\ref{avail}) for each of the Table~\ref{table:Z} categories. 
The horizontal axis in \figref{regret} is the average regret over all users in their $n$th week in the trial, e.g. in their first week, their second week, etc. In the bi-modal setting there are two groups, where all users in group one have a positive response to treatment when experiencing their typical context, while the users in group two have a negative response to treatment under their typical context. An optimal policy would learn to  \textit{not} \textit{typically} send treatments to users in the first group, and to \textit{typically} send them to users in the second. To evaluate each algorithm's ability to learn this distinction we show the percentage of time each group received a message in \tabref{table:bi-modal}.

\begin{table}
\centering
\resizebox{.75\columnwidth}{!}{%
\begin{tabular}{|p{1.75cm}|p{2.7cm }|p{2.7cm }|}
\hline
 & \makecell[l]{\small{Group one} \\ \small{optimal policy} \\ \small{= send activity} \\\small{suggestion} }&  \makecell[l]{\small{Group two} \\ \small{optimal policy} \\ \small{= send anti-sedentary} \\\small{message} }\\
\hline
\small{\complete} &    \multicolumn{1}{r|}{0.49} &   \multicolumn{1}{r|}{0.46}  \\
\hline
\makecell[l]{\small{\textsc{Person-}}\\ \small{\textsc{Specific}} }& \multicolumn{1}{r|}{0.65} &   \multicolumn{1}{r|}{0.49}\\
 \hline
  \makecell[l]{\small{\textsc{GangOB}} }& \multicolumn{1}{r|}{0.57} &   \multicolumn{1}{r|}{0.35}\\
   \hline
\makecell[l]{\small{\textsc{Intelligent-}}\\ \small{\textsc{Pooling} }}& \multicolumn{1}{r|}{0.59} &   \multicolumn{1}{r|}{0.36}     \\
 \hline
\end{tabular}%
}
    \caption{
The fraction of time that  messages were sent to users in each group. 
Recall at each decision time either an activity suggestion or anti-sedentary message is sent. 
For group one it is typically optimal to send an activity suggestion, while for 
group two it is typically optimal to send an anti-sedentary message. Here, \ourapproach{} is 
best able to learn this dynamic. \label{table:bi-modal}}
\end{table}

The relative performance of the approaches depends on the heterogeneity of the population. When the population is
very homogenous \complete{} excels, while its performance suffers as heterogeneity increases.  \none{} is able to personalize; as shown by \tabref{table:bi-modal},
it can differentiate between individuals. However, it  learns slowly and can only approach the performance of \complete{} 
in the smooth setting of Table~\ref{table:Z}
where users differ the most in their response to treatment. Both  \ourapproach{} and  \hobnoisy{} are more adaptive than either \complete{} or \none{}. \hobnoisy{} consistently outperforms \none{}
and achieves lower regret than \complete{} in some settings. In the homeogenous setting we see that \hob{} can utilize social information more effectively than \none{} does while in the smooth setting it can adapt to individual differences more effectively than \complete{}. Yet, \ourapproach{} demonstrates stronger and swifter adaptability than does \hob{}, consistently achieving lower regret at quicker rates. Finally, the algorithms differ in their suitability for real-world applications, especially when data is limited. 
 \hobnoisy{} requires  reliable values for hyper-parameters and can depend on  fixed knowledge about relationships between users.  
 \ourapproach{}   can learn how  to pool between individuals over time and without prior knowledge.

    \section{\ourapproach{}  Feasibility Study}
     \label{sec:clinical}
The simulated experiments provide insights into the potential of this approach for a live deployment. As we see reasonable performance in the simulated setting, we now discuss an initial pilot deployment of \ourapproach{} in a real-life physical activity clinical trial. 

\subsection{Feasibility Study Design}
The feasibility study of \ourapproach{} involves 10 participants added to a larger 90-day clinical trial of HeartSteps v2, an mHealth physical activity intervention.  The purpose of the  larger  clinical trial is to optimize the intervention for individuals with Stage 1 hypertension. 
Study participants with Stage 1 hypertension were recruited from Kaiser Permanente Washington in Seattle, Washington. The study was approved by the institutional review board of the Kaiser Permanente Washington Health Research Institute (under number 1257484-14).

HeartSteps v2 is a cross-platform mHealth application that incorporates several intervention components, including weekly activity goals, feedback on goal progress, planning, motivational messages, prompts to interrupt sedentary behavior, and---most relevant to this paper---actionable, contextually-tailored suggestions for individuals to perform a short physical activity (suggesting, roughly, a 3 to 5 minute walk). In this study physical activity is tracked with a commercial wristband tracker, the Fitbit Versa smart watch.

\begin{figure}
\centering
\begin{minipage}[b]{.95\columnwidth}
    \centering\includegraphics[width=12.0cm]{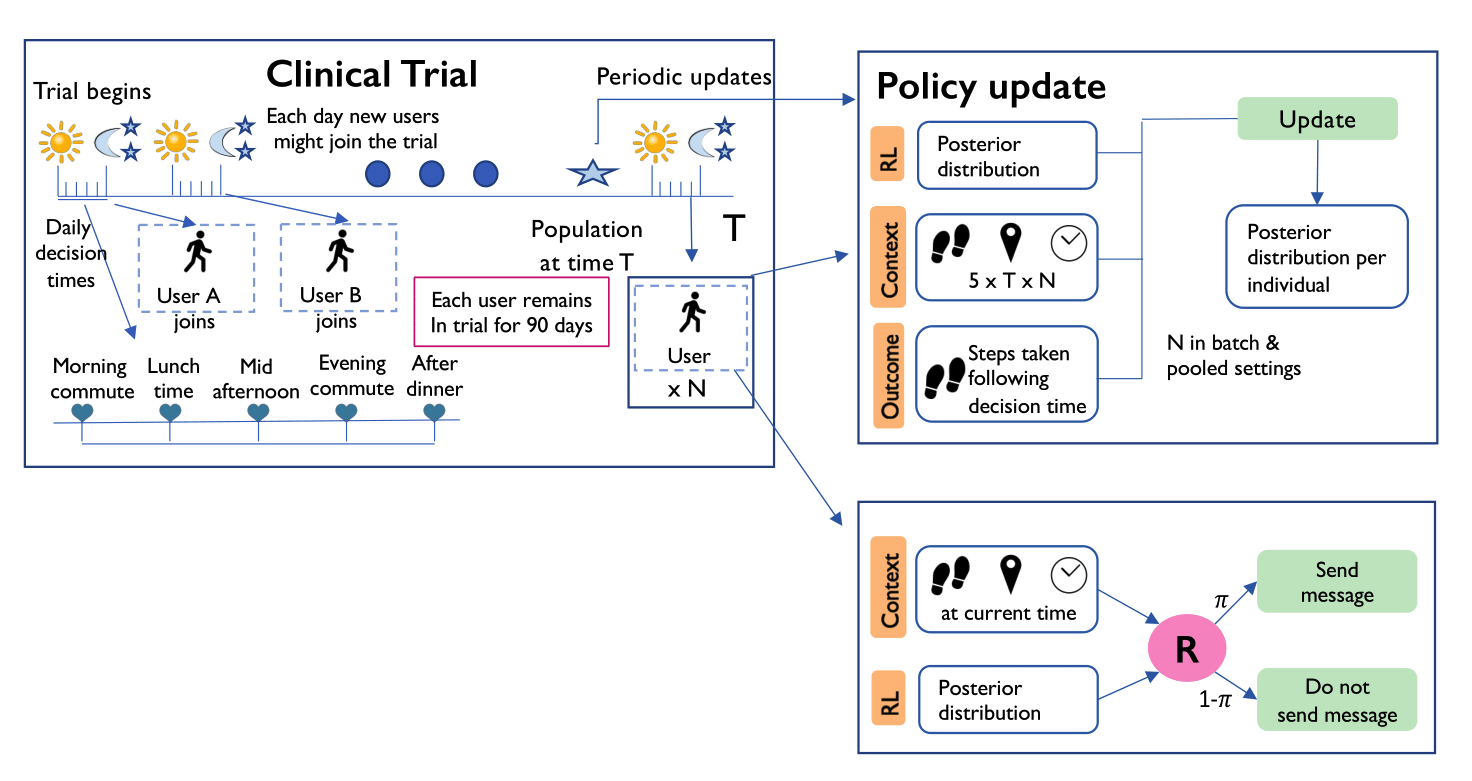}
    \caption{Setup of \clinical{}. Users can receive treatments up to five times a day during the 90 days. Users enter the trial asynchronously. \label{clinical_pic}}
    \end{minipage}
\end{figure}

In this version of the intervention, activity suggestions are randomized  five times per day for each participant on each day of the 90-day trial. These decision times are specified by each user at the start of the study, and they roughly correspond to the participant's typical morning commute, lunch time, mid-afternoon, evening commute, and after dinner periods. The treatment options for activity suggestions are binary: at a decision time, the system can either send or not send a notification with an activity suggestion.  When provided, the content of the suggestion is tailored to current sensor data (location,  weather, time of day, and day of the week). Examples of these suggestions are provided in \cite{klasnja2018}. At a decision time, activity suggestions are randomized only if the system considers that the user is available for the intervention---i.e., that it is appropriate to intervene at that time (see Figure~\ref{available_criteria} for criteria used to determine if it is appropriate to send an activity suggestion at a decision time). Subject to these availability criteria, 
\ourapproach{} determines whether to send a suggestion at each decision time. The posterior distribution was updated once per day, prior to the beginning of each day.  \figref{clinical_pic} provides a schematic of the feasibility study.

The feasibility study included the second set of 10 participants in the trial of HeartSteps v2, following the initial 10 enrolled participants.   \ourapproach{} (\algref{pooledalg}) is deployed  for each of the second set of 10 participants.  At each decision time for these 10 participants, \ourapproach{} uses all data up to that decision time (i.e. from the initial ten participants as well as from the subsequent ten participants).    Thus the feasibility study allows us to assess performance of \ourapproach{} after the beginning of a study instead of the performance at the beginning of the study (when there is little data) or the performance at the end of the study (when there is a large amount of data and the algorithm can be expected to perform well).

In the feasibility study, the features used in the reward model were selected to be predictive of the baseline reward and/or the treatment effect, based on the data analysis of \HSVone; see section 6.2 in \cite{liao2020personalized} for details.  All features used in the reward model are shown in  \tabref{clinical_states}.
The feature \textit{engagement} represents the extent to which a user engages with the mHealth  application measured
as a function of how many screen views are made within the application within a day. 
The feature \textit{dosage} represents the extent to which a user has received treatments (activity suggestions).  This feature  increases and decreases depending on the number of activity suggestions recently received.   The feature \textit{location} refers to whether a user is at home or work (encoded as a 1) or somewhere else (encoded as a 0). The \textit{temperature} feature value is set according to the temperature at a user's current location (based off of phone GPS). The \textit{variation} feature value is set according to the variation in step count in the hour around that decision point over the prior seven-day period. 
As before we construct a feature vector $\phi$, however here we only use select terms to estimate the treatment effect. 
 Here, 
\begin{equation} \label{phi_feature_vector}
\begin{split}
\phi(\state_{i,k},A_{i,k})^T  = & \big( \state_{i,k}^T, \pi_{i,k} \state_{i,k}^{'T}, (A_{i,k}-\pi_{i,k})\state'_{i,k} \big),
\end{split}
\end{equation}
where $\state_{i, k} = \{1, \text{temperature}, \text{yesterday's step count},  \text{preceding} \text{ activity level},   \\ \text{step variation}, \text{engagement}, \text{dosage}, \text{location}\} $
and $\state'_{i, k} = \{1, \text{step variation}, \\ \text{engagement},  \text{dosage}, \text{location}\}$ is a subset of $S_{i, k}$.

We provide a full description of these features in \secref{feature_clinical}. The prior distribution was also constructed based on \HSVone; see Section 6.3 in  \cite{liao2020personalized} for more details. 
As this feasibility study only includes a small number of users, a simple model with only two person-specific random effects, each on the intercept term  in $\state$ and $\state'$ (\eqnref{phi_feature_vector}) was deployed.

\begin{figure}
  \lineskip=-\fboxrule
  \fbox{\begin{minipage}{\dimexpr \textwidth-2\fboxsep-2\fboxrule}
   
A user is available to receive an activity suggestion under the following conditions: 
 \begin{itemize}
 \item She is not currently active and has not had a large amount of activity in the last two hours. 
 \item She has not recently received a notification with a HeartSteps intervention.
 \item Her phone has an internet connection and can communicate with the HeartSteps server.
 \item Her smart watch has been able to communicate with the HeartSteps server in the last ten minutes to provide the current location and step count data.
 \end{itemize}
    \abovecaptionskip=0pt
    \caption{Availability criteria\label{available_criteria}}
  \end{minipage}}
\end{figure}

\begin{table}
\small
\centering
\resizebox{1.0\columnwidth}{!}{%
\begin{tabular}{|p{4.0cm}p{4.0cm}p{1.5cm}p{1.5cm}|}
\hline
    \multicolumn{4}{|c|}{ 
\normalsize{ \textbf{State Features}}} \\
\hline
    \multicolumn{1}{|l|}{\textbf{Name}}  &  \multicolumn{1}{l|}{\textbf{Value} }&  \multicolumn{1}{l|}{\textbf{ \makecell[l]{ \textbf{\patient}\\ Specific}  }}&  \multicolumn{1}{l|}{\textbf{ \makecell[l]{ \textbf{Included in }\\ treatment \\effect}  }}\\
\hline
\hline
Temperature& Continuous & Yes  & No \\
\hline
Yesterday's step count& Continuous & Yes  & No \\
\hline
 Prior 30-minute step count& Continuous& Yes  & No \\
Step variation level& Discrete& Yes & Yes \\
 Engagement with mobile application & Discrete& Yes & Yes  \\
Dosage & Continuous& Yes & Yes  \\
Location & Discrete& Yes  & Yes \\
Intercept & 1& Yes&Yes \\
\hline
    \multicolumn{4}{|c|}{
\normalsize{\textbf{Reward}} } \\
    \hline
Step count & Continuous on log scale& Yes & NA\\
\hline
\end{tabular}
}
\caption{\textmd{State feature descriptions for \clinical.}
\label{clinical_states}}
\end{table}

Here we discuss  how much data we have  to personalize the policy to each user. 
Recall the 10 users only receive interventions when they meet the availability criteria outlined in \figref{available_criteria},
thus we find that in practice we have a limited number of decision points to learn a personalized policy from. 
In the case of perfect availability, we would have at most 450 decision points per person. However due to the criteria in \figref{available_criteria}, the algorithm is used with only  approximately 23\% of each user's decision points. Pooling  users' data
allows us to learn more rapidly. On  the day that the first pooled user joined the feasibility study there were 107 data points from the first set of 10 users.

\begin{figure}
\centering
\begin{minipage}[b]{.95\columnwidth}
    \centering\includegraphics[width=9.0cm]{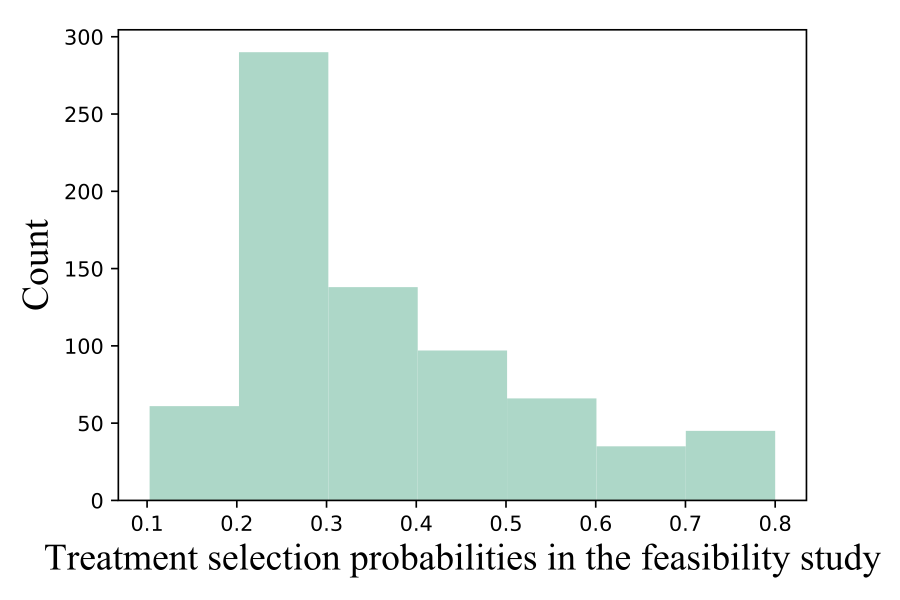}
    \caption{We see that \ourapproach{}  covers the full range of treatment selection probabilities. The tendency seems to be to send with a lower rather than higher probability. \label{rand_hist}}
    \end{minipage}
\end{figure}

The 10 users received an average number of .20 ($\rpm 0.015$) messages a day. The average log step count in the 30-minute window after a suggestion was sent was 4.47, while it was 3.65 in the 30-minute windows after suggestions were not sent. \figref{rand_hist} shows the entire history of treatment selection probabilities for all of the users who received treatment according to \ourapproach{}. We see that the treatment probabilities tended to be low, though they covered the whole range of possible values.

\begin{figure}[h!]
\centering

\begin{minipage}{0.95\linewidth}
    \centering\includegraphics[width=8.0cm]{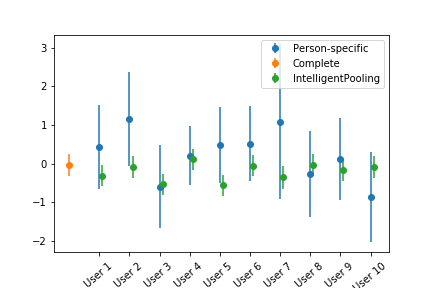}
    \caption{
    Posterior mean and standard deviation of the coefficient of $A_{i,k}$  
    in \eqnref{phi_feature_vector}
for all users in the feasibility study.}\label{all_intercepts}
    \end{minipage}
  \end{figure}
  
  \begin{figure}[h!]
\centering

\begin{minipage}{0.95\linewidth}
    \centering\includegraphics[width=8.0cm]{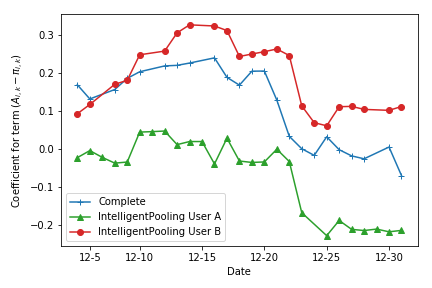}
    \caption{
    Posterior mean of the coefficient of $A_{i,k}$ 
    in \eqnref{phi_feature_vector}
for users A and B in the feasibility study.}\label{two_users}
    \end{minipage}
  \end{figure}

\begin{figure}[h!]
\centering

\begin{minipage}{0.95\linewidth}
    \centering\includegraphics[width=8.0cm]{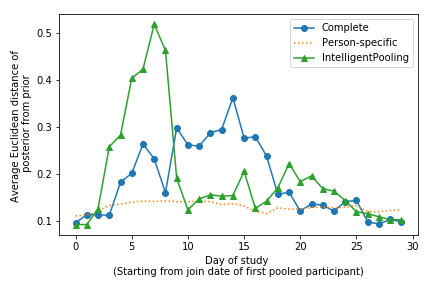}
    \caption{
Mean squared distance of the posterior mean from prior mean of the  coefficients of $A_{i,k}$
} \label{speed}
    \end{minipage}
  \end{figure}

We would like to assess the ability of \ourapproach{} to personalize 
and learn quickly. To do so we perform an analysis of the learning 
algorithms of \ourapproach, \complete{} and \none{} on 
batch data containing tuples of $(S,A,R)$. Note that the actions in this batch data 
were selected by \ourapproach, however, 
here we are not interested in the action selection components of each algorithm 
but instead on their ability to learn the posterior distribution of the weights 
on the feature vector.

\textbf{Personalization}
By comparing how the decisions to treat under \ourapproach{} differ from those under \complete{}, 
 we gather preliminary
 evidence  concerning whether \ourapproach{} personalizes to users. \figref{all_intercepts}
 shows the posterior mean of  the coefficient of the $A_{i,k}$
 term in   the estimation of the treatment effect, for all users in the feasibility study on the 90th day after the last user joined the study. 
 We show this term not only for \ourapproach{} but also for \complete{} and \none. We see that for some users this coefficient is below zero while for others it is above. While the terms under \ourapproach{} differ from \complete{} they do not vary as much as those learned by \none{}. Yet, crucially, the variance is much lower for these terms.

 \figref{two_users} displays the posterior mean of  the coefficient of the $A_{i,k}$ 
 term in  the estimation of the treatment effect.  This coefficient represents 
the overall effect of treatment on one of the  users,  \textit{ User A}.
During the prior 7 days  User A had not experienced much variation in activity at this time and the user's engagement is low. 
Note that the treatment appears to have a positive effect on a different user, User B, in this context whereas on User A there is little evidence of a positive effect.   If \complete{} had been used to determine treatment, User A  might have been over-treated.

\textbf{Speed of policy learning}
We consider the speed at which \ourapproach{} diverges from the prior, relative to the speed of divergence for \none.   \figref{speed} provides the Euclidean distance between the learned posterior and prior parameter vectors (averaged across the data from the 10 users at each time). 
From \figref{speed} we see that  
 \none{} hardly varies over time in contrast to \ourapproach{} and \complete{}, which suggests that \none{} learns more slowly.

In conclusion \ourapproach{} was found to be feasible in this study.  In particular the algorithm was operationally stable within the computational environment of the study, produced decision probabilities in a timely manner, and did not adversely impact the functioning of the overall mHealth intervention application.  Overall, \ourapproach{} produced treatment selection probabilities which covered the full range of available probabilities, though treatments tended to be sent with a low probability.

\section{Non-stationary environments}
\label{sec:time}
An additional challenge in mHealth settings is that users' response to treatment can vary over time. To address this challenge we show 
 that our underlying model can be extended to include time-varying random effects. 
This allows each policy to be aware of how a user's response to treatment might vary over time. 
We propose a new simulation to evaluate this approach and show that \ourapproach{}
achieves state-of-the-art regret, adjusting to non-stationarity
 even as user populations vary from heterogenous to homogenous.

\subsection{Time-varying random effect}
In addition to user-specific random effects we extend our model to include time-specific random effects. Consider the Bayesian mixed effects model with person-specific and time-varying effects: for user $i$ at the $k^{th}$ decision time,
\begin{eqnarray}
{
R_{i, k} = \phi(S_{i, k}, A_{i, k})^\transpose {\weightvector_{i, k}} + \epsilon_{i, k}}.
\label{bayes_reg_time}
\end{eqnarray}
In addition, we impose the following additive structure on the parameters $\weightvector_{i,k}$:
\begin{eqnarray}
\small{
\weightvector_{i,k} = \weightvector_{pop} + u_i \label{random_effect} + v_k
},
\label{randomeffect}
\end{eqnarray}
where $\weightvector_{pop}$ is the population-level parameter, $u_i$
represents the person-specific deviation from $\weightvector_{pop}$ for user $i$ and $v_k$ is the time-varying random effects allowing $w_{i, k}$ to vary with time in the study.

The prior terms for this model are as introduced in \secref{sec:pooling_method}. Additionally, 
$v_k$ has mean $\mathbf{0}$ and covariance $ D_v$. The covariance between two relative decision times in the trial is $\Cov(v_k, v_{k'}) = \rho(k, k') D_v$, where $\rho(k, k') = \exp(-dist(k, k')^2/\sigma_{\rho})$  for a distance function, $dist$ and $\theta_{pop} \independent \{u_i\}  \{v_k\}$. 
There is no change to \algref{pooledalg} except that now the algorithm would select the action based on the posterior distribution of $w_{i, k}$, which depends on both the user and time in the study.

\subsection{Experiments}
We now modify our original simulation environment so that users' responses will vary over time. To do so we introduce the generative model \textbf{\burden{}}. This generative model captures the phenomenon of disengagement. That is as users are increasingly exposed to 
treatment over time they can become less responsive. 
This model adds a further term to (\ref{avail}), $A_{i,k}X_w^T\beta_w   $ where $X_w$ is defined as follows.  Let $w_{i,k}$ be the highest number of weeks user $i$ has completed at time $k$; $X_w$ encodes  a user's current week in a trial, $X_w =[\indicator{w_{i,k}=0},\dots,\indicator{w_{i,k}=11}]$.  We set $\beta_w$ such that the longer a user has been in treatment, the less they respond to  a treatment message. 
When a simulated user is at a decision time the user will receive a treatment message according to whichever RL policy is being run through the simulation.

 In order to evaluate the effectiveness of our time-varying model we compare to Time-Varying Gaussian Process Thompson Sampling  (\timecomp) \cite{bogunovic2016time}. This approach incorporates temporal information for non-stationary environments and was shown to be competitive to stationary models. To compare this method to \ourapproach{} we use a linear kernel for the spatial component. We then modify  \eqnref{post cal} to compute the posterior distribution by removing the random-effects and modifying the kernel (\eqnref{our_kernel}) to include the temporal terms introduced in  \cite{bogunovic2016time}. 
\begin{figure}
  \begin{minipage}[b]{0.95\linewidth}
        \centering\includegraphics[width=9.5cm]{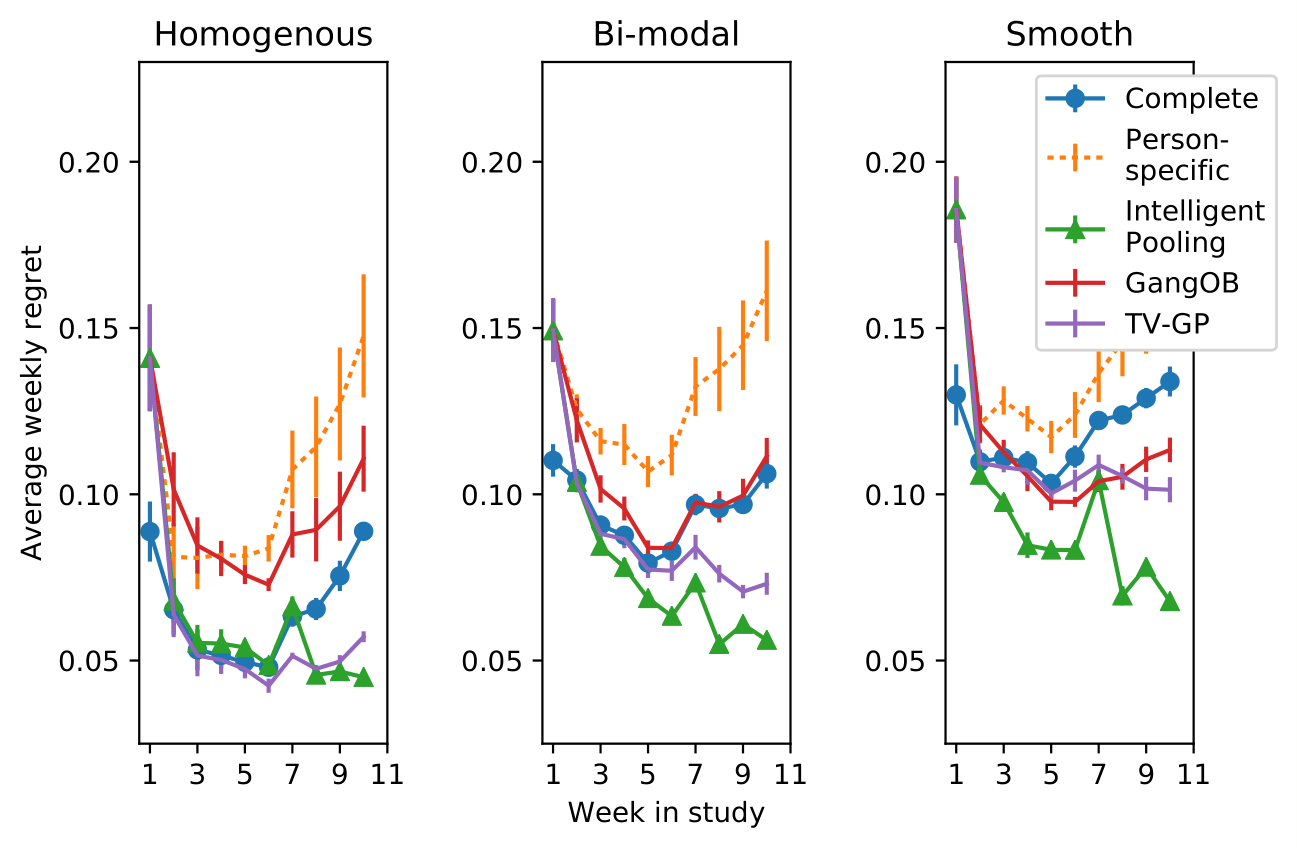}
    \caption{\textbf{\burden{} generative model} Regret averaged across all users for each week in the trial, i.e. average regret of all users in their first week of the trial.}\label{regret_time}
        \end{minipage}
  \end{figure}
  
  \begin{table}[H]
\centering
\resizebox{.75\columnwidth}{!}{%
\begin{tabular}{|p{1.75cm}|p{2.0cm }|p{2.0cm }|}
\hline
 & \makecell[l]{\small{Cohort One} \\ \small{Week 10}}& \makecell[l]{\small{Cohort Six}\\ \small{Week 10}}  \\
\hline
\small{\complete} &    \multicolumn{1}{r|}{0.62} &   \multicolumn{1}{r|}{0.44}  \\
\hline
\makecell[l]{\small{\textsc{Person-}}\\ \small{\textsc{Specific}} }& \multicolumn{1}{r|}{0.76} &   \multicolumn{1}{r|}{0.59}\\
 \hline
  \makecell[l]{\small{\textsc{HordeOB-}} }& \multicolumn{1}{r|}{0.50} &   \multicolumn{1}{r|}{0.57}\\
 \hline
   \makecell[l]{\small{\textsc{TV-GP}} }& \multicolumn{1}{r|}{0.64} &   \multicolumn{1}{r|}{0.31}\\
 \hline
\makecell[l]{\small{\textsc{Intelligent-}}\\ \small{\textsc{Pooling} }}& \multicolumn{1}{r|}{0.30} &   \multicolumn{1}{r|}{0.06}     \\
 \hline
\end{tabular}

}
\caption{Average fraction of times treatment was sent (action=1), over 50 simulations (generative model \populationgen{} with homogenous  $Z^h$ setting). } \label{table:lastweek}
\end{table}

  \figref{regret_time}   provides the regret  averaged across all users across 50 simulated trials where the reward distribution follows generative model $\burden$.   As before the  horizontal axis in \figref{regret_time} is the average regret over all users in their $n^{th}$week in the trial, e.g. in their first week, their second week, etc. In \burden{}, the time-specific response to treatment is set so that a negative response to treatment is introduced in the seventh week of the trial. 

In the \burden{} condition as users become increasingly less responsive to treatment good policies should learn to treat less. Thus, \tabref{table:lastweek} provides  the average number of times a treatment is sent in the last week of the trial for both the first and last cohort. We expect that a policy which learns not to treat will treat less often in the last week of the last cohort than in the last week of the first cohort. 

\section{Limitations}
\label{sec:limitations}
A significant limitation with this work is that our pilot study involved a small number of participants. 
Our results from this work must be considered with caution as preliminary evidence towards the feasibility of
deploying \ourapproach{}, and bandit algorithms in general, in mHealth settings. Moreover, we cannot 
claim to provide generalizable evidence that this algorithm can improve health outcomes; for this larger studies 
with more participants must be run. We offer our findings as motivation for such future work.

Our proposed model is designed to overcome the challenges faced when learning personalized policies in limited data settings. 
As such, if data was abundant our model would likely have limited effectiveness compared to more complex models. 
For example, a more complex model could allow us to pool between users as a function of their similarity. Our current model instead 
determines the extent to which a given user deviates from the population and does not consider between-user similarities. 
A limitation with our current understanding of mHealth is that it is unclear what a good similarity measure would be. We leave the question 
of designing a data-efficient  algorithm for learning such a measure  as future work. 

A component of \ourapproach{} is the use of empirical Bayes to update the model hyper-parameters. Here, we used an approximate procedure. 
However, with our model it is possible to produce exact updates in a streaming fashion and we are currently developing such an approach.

Ideally, we would evaluate \ourapproach{} against all other approaches in a clinical trial setting. 
However, here we only demonstrated the feasibility of our approach on a limited number of users 
and did not have the resources to similarly test the other approaches. 
To overcome this limitation we constructed a realistic simulation environment so that we 
could evaluate on different populations without the costly investment of designing multiple
arms of a real-life trial. While the simulated experiments and the feasibility study together demonstrate
the practicality of our approach, in future work one might deploy all potential approaches in simultaneous live trials.

Finally, \ourapproach{} can incorporate a time-specific random effect to capture the phenomenon of responsivity changing over the course 
of a study. There is much to be improved with this model. For example, the first cohort in a study will not have prior cohorts to learn from, and the final cohort will have the greatest amount of data to benefit from. 
Other models might treat different cohorts with greater equality. 
Furthermore, this representation does not incorporate alternative temporal information, such as continually shifting weather patterns, 
where temperatures might change slowly and gradually alter one's desire to exercise outside. 

\section{Conclusion}
When data on individuals is limited a natural tension exists between personalizing (a choice which can introduce  variance)
and pooling (a choice which can introduce bias).
In this work we have introduced a novel algorithm for personalized
reinforcement learning, \ourapproach{} that presents a principled mechanism for balancing this tension.
 We demonstrate the practicality of our approach in the setting of mHealth.
 In simulation we achieve improvements of 26\% over a state-of-the-art-method, while in a live clinical trial 
we show that our approach shows promise of personalization on even a limited number of users.  
We view adaptive pooling as a first step in addressing the trade-offs between personalization and pooling.
 The question of how to quantify the benefits and risks for individual users is an open direction for future work.

\section*{Acknowledgements}This material is based upon work supported by: 
NIH/NIAAA  R01AA23187 ,NIH/NIDA  P50DA039838,NIH/NIBIB  U54EB020404 and NIH/NCI  U01CA229437.
The views expressed in this article are those of
the authors and do not necessarily reflect the official
position of 
the National Institutes of Health, or any other part of
the U.S. Department of Health and Human Services. 

\section*{Institutional Review Board Approval}
The HeartSteps study discussed here
was approved by the Kaiser Permanente Washington Region Institutional Review Board
under IRB number 1257484-14.

\bibliography{pooling}

\begin{thebibliography}{63}
\providecommand{\natexlab}[1]{#1}
\providecommand{\url}[1]{{#1}}
\providecommand{\urlprefix}{URL }
\expandafter\ifx\csname urlstyle\endcsname\relax
  \providecommand{\doi}[1]{DOI~\discretionary{}{}{}#1}\else
  \providecommand{\doi}{DOI~\discretionary{}{}{}\begingroup
  \urlstyle{rm}\Url}\fi
\providecommand{\eprint}[2][]{\url{#2}}

\bibitem[{Abeille et~al.(2017)Abeille, Lazaric et~al.}]{abeille2017linear}
Abeille M, Lazaric A, et~al. (2017) Linear thompson sampling revisited.
  Electronic Journal of Statistics 11(2):5165--5197

\bibitem[{Agrawal and Goyal(2012)}]{agrawal2012analysis}
Agrawal S, Goyal N (2012) Analysis of thompson sampling for the multi-armed
  bandit problem. In: Conference on Learning Theory, pp 39--1

\bibitem[{Agrawal and Goyal(2013)}]{agrawal2013thompson}
Agrawal S, Goyal N (2013) Thompson sampling for contextual bandits with linear
  payoffs. In: International Conference on Machine Learning, pp 127--135

\bibitem[{Pedregosa~et al.(2011)}]{scikit-learn}
Pedregosa~et al F (2011) Scikit-learn: Machine learning in {P}ython. Journal of
  Machine Learning Research 12:2825--2830

\bibitem[{Bogunovic et~al.(2016)Bogunovic, Scarlett, and
  Cevher}]{bogunovic2016time}
Bogunovic I, Scarlett J, Cevher V (2016) Time-varying {G}aussian process bandit
  optimization. In: Artificial Intelligence and Statistics, pp 314--323

\bibitem[{Bonilla et~al.(2008)Bonilla, Chai, and Williams}]{bonilla2008multi}
Bonilla EV, Chai KM, Williams C (2008) Multi-task {G}aussian process
  prediction. In: Advances in neural information processing systems, pp
  153--160

\bibitem[{Boruvka et~al.(2018)Boruvka, Almirall, Witkiewitz, and
  Murphy}]{boruvka2018assessing}
Boruvka A, Almirall D, Witkiewitz K, Murphy SA (2018) Assessing time-varying
  causal effect moderation in mobile health. Journal of the American
  Statistical Association 113(523):1112--1121

\bibitem[{Brochu et~al.(2010)Brochu, Hoffman, and
  de~Freitas}]{brochu2010portfolio}
Brochu E, Hoffman MW, de~Freitas N (2010) Portfolio allocation for {B}ayesian
  optimization. arXiv preprint arXiv:10095419

\bibitem[{Carlin and Louis(2010)}]{carlin2010bayes}
Carlin BP, Louis TA (2010) Bayes and empirical {B}ayes methods for data
  analysis. Chapman and Hall/CRC

\bibitem[{Casella(1985)}]{casella1985introduction}
Casella G (1985) An introduction to empirical {B}ayes data analysis. The
  American Statistician 39(2):83--87

\bibitem[{Cesa-Bianchi et~al.(2013)Cesa-Bianchi, Gentile, and
  Zappella}]{cesa2013gang}
Cesa-Bianchi N, Gentile C, Zappella G (2013) A gang of bandits. In: Advances in
  Neural Information Processing Systems, pp 737--745

\bibitem[{Cheung et~al.(2018)Cheung, Simchi-Levi, and Zhu}]{cheung2018learning}
Cheung WC, Simchi-Levi D, Zhu R (2018) Learning to optimize under
  non-stationarity. arXiv preprint arXiv:181003024

\bibitem[{Chowdhury and Gopalan(2017)}]{chowdhury2017kernelized}
Chowdhury SR, Gopalan A (2017) On kernelized multi-armed bandits. In:
  International Conference on Machine Learning, vol~70, pp 844--853

\bibitem[{Clarke et~al.(2017)Clarke, Jaimes, and Labrador}]{clarke2017mstress}
Clarke S, Jaimes LG, Labrador MA (2017) mstress: A mobile recommender system
  for just-in-time interventions for stress. In: Consumer Communications \&
  Networking Conference, pp 1--5

\bibitem[{Consolvo et~al.(2008)Consolvo, McDonald, Toscos, Chen, Froehlich,
  Harrison, Klasnja, LaMarca, LeGrand, Libby et~al.}]{consolvo2008activity}
Consolvo S, McDonald DW, Toscos T, Chen MY, Froehlich J, Harrison B, Klasnja P,
  LaMarca A, LeGrand L, Libby R, et~al. (2008) Activity sensing in the wild: a
  field trial of ubifit garden. In: Proceedings of the SIGCHI conference on
  human factors in computing systems, pp 1797--1806

\bibitem[{Desautels et~al.(2014)Desautels, Krause, and
  Burdick}]{desautels2014parallelizing}
Desautels T, Krause A, Burdick JW (2014) Parallelizing exploration-exploitation
  tradeoffs in {G}aussian process bandit optimization. The Journal of Machine
  Learning Research 15(1):3873--3923

\bibitem[{Deshmukh et~al.(2017)Deshmukh, Dogan, and Scott}]{deshmukh2017multi}
Deshmukh AA, Dogan U, Scott C (2017) Multi-task learning for contextual
  bandits. In: Advances in Neural Information Processing Systems, pp 4848--4856

\bibitem[{Djolonga et~al.(2013)Djolonga, Krause, and Cevher}]{djolonga2013high}
Djolonga J, Krause A, Cevher V (2013) High-dimensional {g}aussian process
  bandits. In: Advances in Neural Information Processing Systems, pp 1025--1033

\bibitem[{Finn et~al.(2018)Finn, Xu, and Levine}]{finn2018probabilistic}
Finn C, Xu K, Levine S (2018) Probabilistic model-agnostic meta-learning. In:
  Advances in Neural Information Processing Systems, pp 9516--9527

\bibitem[{Finn et~al.(2019)Finn, Rajeswaran, Kakade, and
  Levine}]{finn2019online}
Finn C, Rajeswaran A, Kakade S, Levine S (2019) Online meta-learning. arXiv
  preprint arXiv:190208438

\bibitem[{Forman et~al.(2018)Forman, Kerrigan, Butryn, Juarascio, Manasse,
  Onta{\~n}{\'o}n, Dallal, Crochiere, and Moskow}]{forman2018can}
Forman EM, Kerrigan SG, Butryn ML, Juarascio AS, Manasse SM, Onta{\~n}{\'o}n S,
  Dallal DH, Crochiere RJ, Moskow D (2018) Can the artificial intelligence
  technique of reinforcement learning use continuously-monitored digital data
  to optimize treatment for weight loss? Journal of behavioral medicine
  42(2):276--290

\bibitem[{Gardner et~al.(2018)Gardner, Pleiss, Weinberger, Bindel, and
  Wilson}]{gardner2018gpytorch}
Gardner J, Pleiss G, Weinberger KQ, Bindel D, Wilson AG (2018) Gpytorch:
  Blackbox matrix-matrix {g}aussian process inference with gpu acceleration.
  In: Advances in Neural Information Processing Systems, pp 7576--7586

\bibitem[{Greenewald et~al.(2017)Greenewald, Tewari, Murphy, and
  Klasnja}]{greenewald2017action}
Greenewald K, Tewari A, Murphy S, Klasnja P (2017) Action centered contextual
  bandits. In: Advances in neural information processing systems, pp 5977--5985

\bibitem[{Gupta et~al.(2018)Gupta, Mendonca, Liu, Abbeel, and
  Levine}]{gupta2018meta}
Gupta A, Mendonca R, Liu Y, Abbeel P, Levine S (2018) Meta-reinforcement
  learning of structured exploration strategies. In: Advances in Neural
  Information Processing Systems, pp 5302--5311

\bibitem[{Hamine et~al.(2015)Hamine, Gerth-Guyette, Faulx, Green, and
  Ginsburg}]{hamine2015impact}
Hamine S, Gerth-Guyette E, Faulx D, Green BB, Ginsburg AS (2015) Impact of
  mhealth chronic disease management on treatment adherence and patient
  outcomes: a systematic review. Journal of medical Internet research 17(2):e52

\bibitem[{Jaimes et~al.(2016)Jaimes, Llofriu, and Raij}]{jaimes2016preventer}
Jaimes LG, Llofriu M, Raij A (2016) Preventer, a selection mechanism for
  just-in-time preventive interventions. IEEE Transactions on Affective
  Computing 7(3):243--257

\bibitem[{Kim and Tewari(2019)}]{kim2019near}
Kim B, Tewari A (2019) Near-optimal oracle-efficient algorithms for stationary
  and non-stationary stochastic linear bandits. arXiv preprint arXiv:191205695

\bibitem[{Kim and Tewari(2020)}]{kim2020randomized}
Kim B, Tewari A (2020) Randomized exploration for non-stationary stochastic
  linear bandits. In: Conference on Uncertainty in Artificial Intelligence, pp
  71--80

\bibitem[{Klasnja et~al.(2015)Klasnja, Hekler, Shiffman, Boruvka, Almirall,
  Tewari, and Murphy}]{klasnja2015microrandomized}
Klasnja P, Hekler EB, Shiffman S, Boruvka A, Almirall D, Tewari A, Murphy SA
  (2015) Microrandomized trials: An experimental design for developing
  just-in-time adaptive interventions. Health Psychology 34(S):1220

\bibitem[{Klasnja et~al.(2018)Klasnja, Smith, Seewald, Lee, Hall, Luers,
  Hekler, and Murphy}]{klasnja2018}
Klasnja P, Smith S, Seewald NJ, Lee A, Hall K, Luers B, Hekler EB, Murphy SA
  (2018) {Efficacy of Contextually Tailored Suggestions for Physical Activity:
  A Micro-randomized Optimization Trial of HeartSteps}. Annals of Behavioral
  Medicine 53(6):573--582

\bibitem[{Krause and Ong(2011)}]{krause2011contextual}
Krause A, Ong CS (2011) Contextual {g}aussian process bandit optimization. In:
  Advances in Neural Information Processing Systems, pp 2447--2455

\bibitem[{Laird et~al.(1982)Laird, Ware et~al.}]{laird1982random}
Laird NM, Ware JH, et~al. (1982) Random-effects models for longitudinal data.
  Biometrics 38(4):963--974

\bibitem[{Lawrence and Platt(2004)}]{lawrence2004learning}
Lawrence ND, Platt JC (2004) Learning to learn with the informative vector
  machine. In: International conference on Machine learning, p~65

\bibitem[{Li et~al.(2010)Li, Chu, Langford, and Schapire}]{li2010contextual}
Li L, Chu W, Langford J, Schapire RE (2010) A contextual-bandit approach to
  personalized news article recommendation. In: Proceedings of the Conference
  on World wide web, pp 661--670

\bibitem[{Li and Kar(2015)}]{li2015context}
Li S, Kar P (2015) Context-aware bandits. arXiv preprint arXiv:151003164

\bibitem[{Liao et~al.(2016)Liao, Klasnja, Tewari, and Murphy}]{liao2016sample}
Liao P, Klasnja P, Tewari A, Murphy SA (2016) Sample size calculations for
  micro-randomized trials in mhealth. Statistics in medicine 35(12):1944--1971

\bibitem[{Liao et~al.(2020)Liao, Greenewald, Klasnja, and
  Murphy}]{liao2020personalized}
Liao P, Greenewald K, Klasnja P, Murphy S (2020) Personalized heartsteps: A
  reinforcement learning algorithm for optimizing physical activity.
  Proceedings of the Conference on Interactive, Mobile, Wearable and Ubiquitous
  Technologies 4(1):1--22

\bibitem[{Luo et~al.(2018)Luo, Yao, Gao, and Zhao}]{luo2018mixed}
Luo L, Yao Y, Gao F, Zhao C (2018) Mixed-effects {G}aussian process modeling
  approach with application in injection molding processes. Journal of Process
  Control 62:37--43

\bibitem[{Morris(1983)}]{morris1983parametric}
Morris CN (1983) Parametric empirical {B}ayes inference: theory and
  applications. Journal of the American statistical Association 78(381):47--55

\bibitem[{Nagabandi et~al.(2018)Nagabandi, Finn, and
  Levine}]{nagabandi2018deep}
Nagabandi A, Finn C, Levine S (2018) Deep online learning via meta-learning:
  Continual adaptation for model-based rl. arXiv preprint arXiv:181207671

\bibitem[{Nahum-Shani et~al.(2017)Nahum-Shani, Smith, Spring, Collins,
  Witkiewitz, Tewari, and Murphy}]{nahum2017just}
Nahum-Shani I, Smith SN, Spring BJ, Collins LM, Witkiewitz K, Tewari A, Murphy
  SA (2017) Just-in-time adaptive interventions ({J}{I}{T}{A}{I}s) in mobile
  health: key components and design principles for ongoing health behavior
  support. Annals of Behavioral Medicine 52(6)

\bibitem[{Paredes et~al.(2014)Paredes, Gilad-Bachrach, Czerwinski, Roseway,
  Rowan, and Hernandez}]{paredes2014poptherapy}
Paredes P, Gilad-Bachrach R, Czerwinski M, Roseway A, Rowan K, Hernandez J
  (2014) Poptherapy: Coping with stress through pop-culture. In: Conference on
  Pervasive Computing Technologies for Healthcare, pp 109--117

\bibitem[{Qi et~al.(2018)Qi, Wu, Wang, Tang, and Sun}]{NEURIPS2018_d8c9d05e}
Qi Y, Wu Q, Wang H, Tang J, Sun M (2018) Bandit learning with implicit
  feedback. In: Advances in Neural Information Processing Systems, vol~31, pp
  7276--7286

\bibitem[{Qian et~al.(2019)Qian, Klasnja, and Murphy}]{qian2019linear}
Qian T, Klasnja P, Murphy SA (2019) Linear mixed models under endogeneity:
  modeling sequential treatment effects with application to a mobile health
  study. arXiv preprint arXiv:190210861

\bibitem[{Rabbi et~al.(2015)Rabbi, Aung, Zhang, and
  Choudhury}]{rabbi2015mybehavior}
Rabbi M, Aung MH, Zhang M, Choudhury T (2015) Mybehavior: automatic
  personalized health feedback from user behaviors and preferences using
  smartphones. In: Proceedings of the Conference on Pervasive and Ubiquitous
  Computing, pp 707--718

\bibitem[{Rabbi et~al.(2017)Rabbi, Philyaw-Kotov, Lee, Mansour, Dent, Wang,
  Cunningham, Bonar, Nahum-Shani, Klasnja et~al.}]{rabbi2017sara}
Rabbi M, Philyaw-Kotov M, Lee J, Mansour A, Dent L, Wang X, Cunningham R, Bonar
  E, Nahum-Shani I, Klasnja P, et~al. (2017) S{A}{R}{A}: a mobile app to engage
  users in health data collection. In: Joint Conference on Pervasive and
  Ubiquitous Computing and the International Symposium on Wearable Computers,
  pp 781--789

\bibitem[{Raudenbush and Bryk(2002)}]{raudenbush2002hierarchical}
Raudenbush SW, Bryk AS (2002) Hierarchical linear models: Applications and data
  analysis methods, vol~1

\bibitem[{Russac et~al.(2019)Russac, Vernade, and
  Capp{\'e}}]{russac2019weighted}
Russac Y, Vernade C, Capp{\'e} O (2019) Weighted linear bandits for
  non-stationary environments. In: Advances in Neural Information Processing
  Systems, pp 12017--12026

\bibitem[{Russo and Van~Roy(2014)}]{russo2014learning}
Russo D, Van~Roy B (2014) Learning to optimize via posterior sampling.
  Mathematics of Operations Research 39(4):1221--1243

\bibitem[{Russo et~al.(2018)Russo, Roy, Kazerouni, Osband, and Wen}]{russo2018}
Russo DJ, Roy BV, Kazerouni A, Osband I, Wen Z (2018) A tutorial on thompson
  sampling. Foundations and Trends in Machine Learning 11(1):1--96,
  \urlprefix\url{http://dx.doi.org/10.1561/2200000070}

\bibitem[{S{\ae}mundsson et~al.(2018)S{\ae}mundsson, Hofmann, and
  Deisenroth}]{saemundsson2018meta}
S{\ae}mundsson S, Hofmann K, Deisenroth MP (2018) Meta reinforcement learning
  with latent variable {g}aussian processes. arXiv preprint arXiv:180307551

\bibitem[{Shi et~al.(2012)Shi, Wang, Will, and West}]{shi2012mixed}
Shi J, Wang B, Will E, West R (2012) Mixed-effects {G}aussian process
  functional regression models with application to dose--response curve
  prediction. Statistics in medicine 31(26):3165--3177

\bibitem[{Srinivas et~al.(2009)Srinivas, Krause, Kakade, and
  Seeger}]{srinivas2009gaussian}
Srinivas N, Krause A, Kakade SM, Seeger M (2009) Gaussian process optimization
  in the bandit setting: No regret and experimental design. International
  Conference on Machine Learning p 1015–1022

\bibitem[{Thompson(1933)}]{thompson1933likelihood}
Thompson WR (1933) On the likelihood that one unknown probability exceeds
  another in view of the evidence of two samples. Biometrika 25(3/4):285--294

\bibitem[{Vaswani et~al.(2017)Vaswani, Schmidt, and
  Lakshmanan}]{vaswani2017horde}
Vaswani S, Schmidt M, Lakshmanan L (2017) Horde of bandits using {G}aussian
  {M}arkov random fields. In: Artificial Intelligence and Statistics, pp
  690--699

\bibitem[{Wang and Khardon(2012)}]{wang2012nonparametric}
Wang Y, Khardon R (2012) Nonparametric {B}ayesian mixed-effect model: A sparse
  {G}aussian process approach. arXiv preprint arXiv:12116653

\bibitem[{Wang et~al.(2016)Wang, Zhou, and Jegelka}]{wang2016optimization}
Wang Z, Zhou B, Jegelka S (2016) Optimization as estimation with {G}aussian
  processes in bandit settings. In: Artificial Intelligence and Statistics, pp
  1022--1031

\bibitem[{Williams and Rasmussen(2006)}]{williams2006gaussian}
Williams CK, Rasmussen CE (2006) Gaussian processes for machine learning,
  vol~2. MIT press Cambridge, MA

\bibitem[{Xia(2018)}]{xia2018price}
Xia I (2018) The price of personalization: An application of contextual bandits
  to mobile health. Senior thesis

\bibitem[{Yom-Tov et~al.(2017)Yom-Tov, Feraru, Kozdoba, Mannor, Tennenholtz,
  and Hochberg}]{yom2017encouraging}
Yom-Tov E, Feraru G, Kozdoba M, Mannor S, Tennenholtz M, Hochberg I (2017)
  Encouraging physical activity in patients with diabetes: intervention using a
  reinforcement learning system. Journal of medical Internet research
  19(10):e338

\bibitem[{Zhao et~al.(2020)Zhao, Zhang, Jiang, and Zhou}]{zhao2020}
Zhao P, Zhang L, Jiang Y, Zhou ZH (2020) A simple approach for non-stationary
  linear bandits. In: Proceedings of the Conference on Artificial Intelligence
  and Statistics, pp 746--755

\bibitem[{Zhou et~al.(2018)Zhou, Mintz, Fukuoka, Goldberg, Flowers, Kaminsky,
  Castillejo, and Aswani}]{zhou2018personalizing}
Zhou M, Mintz Y, Fukuoka Y, Goldberg K, Flowers E, Kaminsky P, Castillejo A,
  Aswani A (2018) Personalizing mobile fitness apps using reinforcement
  learning. In: CEUR workshop proceedings, vol 2068

\bibitem[{Zintgraf et~al.(2019)Zintgraf, Shiarlis, Kurin, Hofmann, and
  Whiteson}]{zintgraf2019caml}
Zintgraf LM, Shiarlis K, Kurin V, Hofmann K, Whiteson S (2019) {CAML}: Fast
  context adaptation via meta-learning. In: International Conference on Machine
  Learning, pp 7693--7702

\end{thebibliography}
\bibliographystyle{spbasic}   
\appendix

\newpage
\section{Regret Bound}
\label{sec:proof_details}

In this section we prove a high probability regret bound for a modification of \ourapproach{} in a simplified setting.  We modify the Thompson sampling algorithm in \ourapproach{} by multiplying the posterior covariance by a tuning parameter, following  \cite{agrawal2012analysis}. This is mainly due to the technical reasons; see \cite{abeille2017linear} for a discussion. We also simplify the setting in this regret analysis. Specifically, we assume that the posterior distribution of all users is updated after every decision time and the hyper-parameters are fixed throughout the study.

\citet{vaswani2017horde} also provided a regret bound for the Thompson Sampling Horde of Bandits algorithm where the data is pooled using a known, prespecified, social graph. 
\citet{vaswani2017horde} employ the conceptual framework of  \citet{agrawal2012analysis} 
which uses the concept of \textit{saturated} and \textit{unsaturated} arms 
to bound the regret. They show that the regret for playing an arm from the unsaturated set (which includes the optimal arm)
can be bounded by a factor of the standard deviation which decreases over time. Additionaly, they show that the probability of playing a saturated arm is  small, so that an unsaturated arm will be played at each time with some constant probability. 
\citet{vaswani2017horde} follow this argument, but adapt their proof to include the prior covariance of the social graph 
in the bound of the variance. 
Our proof follows along similar lines with the primary difference being how the prior covariance of all parameters is formulated. Specifically,  the prior variance in \cite{vaswani2017horde} is constructed by the Laplacian matrix of the social graph, whereas ours is constructed based on the Bayesian mixed effects model \eqref{randomeffect1}. As a result, while in  \citet{vaswani2017horde} the regret bound is stated in terms of properties of the social graph, our bound depends on properties of our mixed effects model (i.e., the covariance matrix of the random effects).

Recall that $\Sigma_w$ is the prior covariance of the weight vector $\weightvector_{pop}$, $\Sigma_u$ is the covariance of the random effect $u_i$ and $\sigma_\epsilon^2$ is the variance of the error term. We assume that both $\weightvector_{pop}$ and $u_{i}$ have the same dimensions and that $\Sigma_u$ is invertible. Additionally, for simplicity of presentation we assume that the largest eigenvalue in $\Sigma_w$ is at most $d$ and the largest eigenvalue of $\Sigma_u$ is at most $dN$.

\bigskip
\bigskip

Recall that \theoremref{regret_bound} bounds the regret of \ourapproach{} at time $T$ by:
$$\mathcal{R}(T) = \mathcal{\tilde{O}}\Bigg(dN\sqrt{T}\sqrt{\log\Big(\frac{( \trace{\Sigma_\weightvector}+\trace{\Sigma_u}+\trace{ \Sigma_u^{-1})}}{d}+\frac{T}{\sigma_\epsilon^2dN} \Big) \log{\frac{1}{\delta}}}\Bigg)$$ with probability $1-\delta$.
\bigskip

\noindent \textit{Proof Sketch of \theoremref{regret_bound}}. 
We align the decision times from all users by the calendar time.  Specifically, for a given time $t$, we retrieve the user index encountered at time $t$ by $i(t)$ and retrieve this user's decision time index by $k(t)$. 
\ourapproach{} selects an action $A_{i(t), k(t)} \in \A$ for time $t \in [1,\dots,T]$. We denote the selected action at time $t$ by $\selectedaction_t$. 

In this setting, we combine each user specific variable into a global shared variable. 
Recall that a feature vector $\phi(A_{i,k},S_{i,k})$ encodes contextual variables for the  action and state of user $i$ at their $k^{th}$ decision time. For simplicity, we denote by $\selectedaction_t$ the action $A_{i(t),k(t)}$ at time $t$ and denote the vector $\phi(A_{i(t),k(t)},S_{i(t),k(t)})$ at time $t$ by $\phi_{\selectedaction_t,t}$. {Additionally, we let $\phi_{\action,t}$ refer to 
$\phi(\action,S_{i(t), k(t)})$ for any $\action \in \A$. }
We introduce a sparse vector $\globcv_{\selectedaction_t,t} \in  \mathbb{R}^{dN}$, which contains $\contextvector_{\selectedaction_t,t}$ 
vector among $N$ $d$-dimensional vectors, the rest of which are zeros .

In proving the regret we consider the equivalent way of selecting the action. Instead of randomizing the action by the probability, here to select an action we assume the algorithm draws a sample $\tilde w_t = \tilde w_{i(t),k(t)}$ and then selects the action $\selectedaction_t=A_{i(t), k(t)} = \argmax_{a \in \A} \phi_{a, t}^T \tilde w_t$ that maximizes the sampled reward.  Analogously to $\phi_{\action, t}$, we define $\hat{\globfv}_t$ and $\tilde{\globfv}_t$  as the  sparse vectors  which contain  $\hat{\weightvector}_{i(t),k(t)}$ and $\tilde{\weightvector}_{i(t),k(t)}$ respectively  as the $i(t)$-th vector among $Nd$-dimensional vector, the rest of which are zeros.

We concatenate the person-specific parameters $w_i$ into $\globfv \in \mathbb{R}^{dN}$. Let the prior covariance of $\globfv$ be $\priorcov = \mathbf{1}_{N\times N}\otimes\Sigma_w+\mathbf{I}_{N} \otimes \Sigma_u$. 
 At time $t$, all contexts observed thus far, for all users, can be combined into one matrix $\matcv_t \in \mathbb{R}^{t\times dN}$
 where a single row $s$ corresponds
to $\globcv_{\action_s,s}$,
the sparse context vector associated with the action $A_s$ taken for user $i(s)$ at their $k(s)$-th decision time. 
 Let, $\Omega_t = \frac{1}{\sigma_\epsilon^2}\matcv_t^\transpose\matcv_t +\priorcov$.
  At each decision time $t$ we draw a feature vector   $\tilde{\globfv}_t \sim \mathcal{N}(\hat{\globfv_t},v_t^2\Omega_t^{-1})$.

   Now, within this framework, we rewrite the instantaneous regret as $\reggap=\globcv_{\action^*_t,t}^\transpose \globfv_{t} - \globcv_{A_t,t}^\transpose \globfv_{t}$.
 We prove that with high probability both $\globcv_{\action,t}^\transpose \hat{\globfv_t}$ and $\globcv_{\action,t}^\transpose \tilde{\globfv_t}$ are concentrated around their respective means. 
 The standard deviation around the reward at decision time $t$ for action $\action$ is thus $s_{\action,t}=\sqrt{\globcv_{\action,t}^\transpose\Omega_{t-1}^{-1}\globcv_{\action,t}}$. 
We proceed as in \cite{agrawal2012analysis,vaswani2017horde}  by bounding three terms, the event $\mathcal{E}^{\theta_t}$, the event $\mathcal{E}^{\globfv_t}$ and $\sum_{t=1}^T s_{A_t,t}^2$

\begin{definition}
Let $\eminu$
be the inverse of the smallest eigenvalue of $\Sigma_u$,
$\emaxu$ be the largest eigenvalue of $\Sigma_u$, $\emaxp$ be the largest eigenvalue of $\Sigma_w$
and let $\emaxknot = \emaxu+\emaxp$. We assume that $\emaxu\leq dN$ and $\emaxp\leq d$.
\end{definition}

\begin{definition}
For all $\action$, define $\theta_{\action,t} =\globcv_{\action,t}^\transpose \tilde{\globfv}_{t}$.

\end{definition}

\begin{definition}
\begin{align*}
l_t&=\sqrt{dN\log\Big(1+\frac{\emaxknot\eminu}{\delta}  +\frac{t\eminu}{dN\delta}\Big)}+\sqrt{N\emaxp+\emaxu} \\
v_t&=2\sqrt{dN\log\Big(1+\frac{\emaxknot\eminu}{\delta}  +\frac{t\eminu}{dN\delta}\Big)}\\
g_t&=\text{min}\{\sqrt{4dN\natlog(t)},\sqrt{4\natlog(|\mathcal{A}|t)}\}v_t +l_T.
\end{align*}
\end{definition}

\begin{definition}
Define $\mathcal{E}^{\globfv_t}$ and $\mathcal{E}^{\theta_t}$ as the events that $\globcv_t^\transpose \hat{\globfv_t}$ and $\theta_{A_t,t}$ are concentrated around their respective means. Recall that $|\mathcal{A}|$ is the total number of actions. Formally, define  $\mathcal{E}^{\globfv_t}$  as the event that
$$\forall \action: |\globcv_{\action,t}^\transpose \hat{\globfv_t}-\globcv_{\action,t}^\transpose \globfv| \leq l_t s_{\action,t}.$$
Define $\mathcal{E}^{\theta_{t}}$ as the event that 
$$\forall \action: | \theta_{A_t,t} - \globcv_{A_t, t}^\transpose \hat{\globfv}_t|\leq \text{min}\{4dN\log(t),4\log(|\mathcal{A}|t) \}v_ts_{\action,t}. $$
\end{definition}

Let $\zeta = \frac{1}{4e\sqrt{\pi}}$. Given that the events $\mathcal{E}^{\globfv_t}$ and $\mathcal{E}^{\theta_t}$  hold with high probability, we follow an argument similar to Lemma 4 of \cite{agrawal2012analysis} and obtain the following bound:

\begin{align}
\label{regret_inq}
\mathcal{R}(T)\leq \frac{3g_T}{\zeta}\sum_{t=1}^T s_{A_t,t} + \frac{2g_T}{\zeta}\sum_{t=1}^T \frac{1}{t^2}+6g_T\sqrt{|\mathcal{A}|T\log(2/\delta)}.
\end{align}

To bound the variance of the selected actions, $\sum_{t=1}^T s_{A_t,t}$, we follow an argument similar to \cite{vaswani2017horde}, and include the prior covariance terms of our model.  
We prove the following inequality: 

\begin{align}
\label{std}
\sum_{t=1}^T s_{A_t,t} \leq \sqrt{dNT}\sqrt{C\Big(\log\Big(\frac{( \trace{\Sigma_\weightvector}+\trace{\Sigma_u}+\trace{ \Sigma_u^{-1})}}{d}+\frac{T}{\sigma_\epsilon^2dN}\Big)\Big)},
\end{align}
where $C$ is a constant equal to $\frac{\eminu}{\log(1+\frac{\eminu}{\sigma_\epsilon^2})}$. By combining \eqnref{regret_inq} and \eqnref{std} we obtain the bound given in Theorem \ref{regret_bound}. \hfill$\square$\\

\section{Supporting Lemmas}

\begin{definition}
Recall that at time $t$ we define as $\mathcal{D}_t$ as the history of all observed states, actions, and rewards up to time $t$.
Define filtration $\filtration_{t-1}$ as the union of history until time $t-1$, and the contexts at time t, i.e., $\filtration_{t-1} = \{\mathcal{D}_{t-1},\globcv_{\action,t}, \action \in \A\}.$ 
By definition, $\filtration_{1} \subseteq \filtration_{2} \dots \subseteq \filtration_{t-1}$. The following quantities are also determined by the history $\mathcal{D}_{t-1}$ and the contexts, 
$\globcv_{\action,t}$ and are included in $\filtration_{t-1}$.
\begin{itemize}
\item $\hat{\globfv}_t, \Omega_{t-1}$
\item $s_{\action,t} \forall \action$
\item the identity of the optimal action $\action^*_t$
\item whether $\mathcal{E}^\globfv_t$ is true or not
\item the distribution of $\normal(\hat{\globfv}_t,v_t^2\Omega_{t-1}^{-1})$
\end{itemize}
Note that the actual action $A_t$ which is selected at decision point $t$ is not included in  $\filtration_{t-1}$.
\end{definition}

We now address the lemmas used in the proof which differ from \cite{agrawal2012analysis,vaswani2017horde}.

\begin{lemma}
For $\delta \in (0,1)$ :
$$Pr(\mathcal{E}^{\globfv_t})\geq 1 - \frac{\delta}{2}$$
\end{lemma}

\textit{Proof}
The true reward at time $t$, $R_t = \globcv_{A_t,t}^\transpose \globfv + \epsilon_{t}$. Let, $\Omega_t \hat\globfv_t= \frac{\mathbf{b_t}}{\sigma^2_\epsilon}$. Define $\mathbf{S}_{t-1} = \sum_{l=1}^{t-1} \epsilon_{l}\globcv_{\action_l,l}$.
\begin{align*}
&\mathbf{S}_{t-1} = \sum_{l=1}^{t-1} (R_l -\globcv_{\action_l,l}^\transpose \globfv)\globcv_{\action_l,l}= \sum_{l=1}^{t-1} (R_l\globcv_{\action_l,l} -\globcv_{\action_l,l}\globcv_{\action_l,l}^\transpose \globfv)\\
&\mathbf{S}_{t-1} = b_{t-1}-\sum_{l=1}^{t-1}( \globcv_{\action_l,l}\globcv_{\action_l,l}^\transpose \globfv) =  b_{t-1} - \sigma^2_\epsilon(\Omega_{t-1}\hat\globfv_t - \Omega_{t-1}\globfv+\Sigma_{0}\globfv)\\
&\hat\globfv_t-\globfv = \Omega_{t-1}^{-1}\big(\frac{\mathbf{S}_{t-1}}{\sigma^2_\epsilon}-\Sigma_0\globfv \big).
\end{align*}

The following holds for all $\action$: 

\begin{align*}
|\globcv_{\action,t}^\transpose\hat\globfv_t-\globcv_{\action,t}^\transpose\globfv|&=|\globcv_{\action,t}^\transpose(\hat\globfv_t-\globfv)|\\
&\leq \big| \globcv_{\action,t} \Omega_{t-1}^{-1}\big(\frac{\mathbf{S}_{t-1}}{\sigma^2_\epsilon}-\Sigma_0\globfv \big)\big| \\
&\leq \| \globcv_{\action,t}\|_ {\Omega_{t-1}^{-1}} \Big( \Big\|\frac{\mathbf{S}_{t-1}}{\sigma^2_\epsilon} -\Sigma_0\globfv\Big\|_ {\Omega_{t-1}^{-1}}   \Big).
\end{align*}

By the triangle inequality, 

\begin{equation}
\label{triangle}
|\globcv_{\action,t}^\transpose\hat\globfv_t-\globcv_{\action,t}^\transpose\globfv|\leq  \Big( \Big\|\frac{\mathbf{S}_{t-1}}{\sigma^2_\epsilon}\Big\|_ {\Omega_{t-1}^{-1}}  +\|\Sigma_0\globfv\|_ {\Omega_{t-1}^{-1}}   \Big)
\end{equation}

We now bound the term $\|\Sigma_0\globfv\|_ {\Omega_{t-1}^{-1}} $. Recall that the prior covariance of $\globfv, \priorcov = \mathbf{1}_{N\times N}\otimes\Sigma_w+\mathbf{I}_{N} \otimes \Sigma_u$. 

\begin{align*}
\nu_{\text{max}}(\priorcov) &= \nu_{\text{max}}(\mathbf{1}_{N\times N}\otimes\Sigma_w+\mathbf{I}_{N} \otimes \Sigma_u)\\
&=\nu_{\text{max}}(\mathbf{1}_{N\times N})\cdot \nu_{\text{max}}(\Sigma_w)+\nu_{\text{max}}(\mathbf{I}_{N}) \cdot \nu_{\text{max}}(\Sigma_u)\\
&=N\nu_{\text{max}}(\Sigma_w)+\nu_{\text{max}}(\Sigma_u)\\
&=N\emaxp+\emaxu
\end{align*}

\begin{align*}
\|\Sigma_0\globfv\|_ {\Omega_{t-1}^{-1}}& \leq \|\Sigma_0\globfv\|_ {\Sigma_{0}^{-1}} = \sqrt{\globfv\Sigma_0^\transpose\Sigma_0^{-1}\Sigma_0\globfv}
&= \sqrt{\globfv^\transpose\priorcov\globfv}\\
&\leq \sqrt{\nu_{\text{max}}(\priorcov)\|\globfv\|_2}\\
&\leq \sqrt{\nu_{\text{max}}(\priorcov)}\\
&\leq \sqrt{N\emaxp+\emaxu}
\end{align*}

For bounding $ \| \globcv_{\action,t}\|_ {\Omega_{t-1}^{-1}}$, note that
$$ \| \globcv_{\action,t}\|_ {\Omega_{t-1}^{-1}} = \sqrt{\globcv_{\action,t}^\transpose \Omega_{t-1}^{-1} \globcv_{\action,t}} = s_{\action,t}$$.

We can thus write \eqnref{triangle}
\begin{equation}
\label{trianglespecific}
|\globcv_{\action,t}^\transpose\hat\globfv_t-\globcv_{\action,t}^\transpose\globfv|\leq  s_{\action,t}\Big(\frac{1}{\sigma_\epsilon} \Big\|\mathbf{S}_{t-1}\Big\|_ {\Omega_{t-1}^{-1}}  +  \sqrt{n\emaxp+\emaxu}  \Big)
\end{equation}

We now bound $\Big\|\mathbf{S}_{t-1}\Big\|_ {\Omega_{t-1}^{-1}}$. 

\begin{theorem}
For any $d>0, t\geq1$, with probability at least $1-\delta$, 
\begin{align*}
\Big\|\mathbf{S}_{t-1}\Big\|_ {\Omega_{t-1}^{-1}}^2&\leq2\sigma_\epsilon^2 \log \Big( \frac{\det{\Omega_t}^{\frac{1}{2}}\det{\Sigma_0}^{\frac{-1}{2}}}{\delta}\Big)\\
&\leq 2\sigma_\epsilon^2 \Big( \log(\det{\Omega_t}^{\frac{1}{2}})+\log(\det{\Sigma_0}^{\frac{-1}{2}})-\log(\delta) \Big)\\
&\leq \sigma_\epsilon^2 \Big( \log(\det{\Omega_t})+\log(\det{\Sigma_0}^{-1})-2\log(\delta) \Big).
\end{align*}
\end{theorem}

For any $n\times n $ matrix $A$, $\det(A)\leq \big(\frac{\trace{A}}{n}\big)^n$. This implies, $\log(\det(A))\leq n \log \big( \frac{\trace{A}}{n}\big)$. Applying this inequality for both $\Omega_t$ and $\priorcov^{-1}$, we obtain: 

\begin{equation}
\label{eq:tracedet}
\Big\|\mathbf{S}_{t-1}\Big\|_ {\Omega_{t-1}^{-1}}\leq dN\sigma_\epsilon^2\Big( \log\Big(\frac{\trace{\Omega_t}}{dN}\Big) +\log\Big(\frac{\trace{\Sigma_0^{-1}}}{dN} \Big)-\frac{2}{dN}\log(\delta) \Big)
\end{equation}

Next, we use the fact that 

$$\Omega_t = \priorcov + \Sigma_{l=1}^t \globcv_{\action_l,l}\globcv_{\action_l,l}^\transpose \Rightarrow \trace{\Omega_t}\leq\trace{\priorcov}+t$$

\begin{align*}
\trace{\priorcov} &= \trace{ \mathbf{1}_{N\times N}\otimes\Sigma_w+\mathbf{I}_{N} \otimes \Sigma_u}\\
&=\trace{ \mathbf{1}_{N\times N}}\cdot \trace{\Sigma_w}+\trace{\mathbf{I}_{N}} \cdot \trace{\Sigma_u}\\
&=N \trace{\Sigma_w}+N \trace{\Sigma_u} = N( \trace{\Sigma_w}+ \trace{\Sigma_u})
\end{align*}

We now return to \eqnref{eq:tracedet}
\begin{align*}
\Big\|\mathbf{S}_{t-1}\Big\|_ {\Omega_{t-1}^{-1}}^2&\leq dN\sigma_\epsilon^2\Big( \log\Big(\frac{\trace{\Sigma_0 }+t}{dN}\Big) +\log\Big(\frac{\trace{\Sigma_0^{-1}}}{dN}\Big) -\frac{2}{dN}\log(\delta) \Big)\\
&\leq dN\sigma_\epsilon^2\Big( \log\Big(\frac{\trace{\Sigma_0 }\trace{\Sigma_0^{-1}}+t\trace{\Sigma_0^{-1}}}{d^2N^2}\Big) -\log(\delta^\frac{2}{dN}) \Big)\\
& = dN\sigma_\epsilon^2\Big( \log\Big(\frac{\trace{\Sigma_0 }\trace{\Sigma_0^{-1}}+t\trace{\Sigma_0^{-1}}}{d^2N^2\delta}\Big)  \Big)\\
&\leq dN\sigma_\epsilon^2\Big( \log\Big(\frac{d^2N^2\emaxknot\eminu  +tdN\eminu}{d^2N^2\delta}\Big)  \Big)\\
&=dN\sigma_\epsilon^2\Big( \log\Big(\frac{\emaxknot\eminu}{\delta}  +\frac{t\eminu}{dN\delta}\Big)  \Big)\\
\Big\|\mathbf{S}_{t-1}\Big\|_ {\Omega_{t-1}^{-1}}&\leq \sigma_\epsilon\sqrt{dN\log\Big(\frac{\emaxknot\eminu}{\delta}  +\frac{t\eminu}{dN\delta}\Big)}\\
\Big\|\mathbf{S}_{t-1}\Big\|_ {\Omega_{t-1}^{-1}}&\leq \sigma_\epsilon\sqrt{dN\log\Big(1+\frac{\emaxknot\eminu}{\delta}  +\frac{t\eminu}{dN\delta}\Big)}
\end{align*}

\begin{align*}
|\globcv_{\action,t}^\transpose\hat\globfv_t-\globcv_{\action,t}^\transpose\globfv|&\leq  s_{\action,t}\sqrt{dN\log\Big(1+\frac{\emaxknot\eminu}{\delta}  +\frac{t\eminu}{dN\delta}\Big)}+\sqrt{N\emaxp+\emaxu}\\
&\leq s_{\action,t}l_t
\end{align*}
\hfill$\square$\\

\begin{lemma}
\label{inst:regret}
With probability $1-\frac{\delta}{2}$, 
\begin{equation}
\sum_{t=1}^T \text{regret}(t)\leq \sum_{t=1}^T \frac{3g_t}{\zeta}s_t+ \sum_{t=1}^T \frac{2g_t}{\zeta t^2}s_t+ \sqrt{2\sum_{t=1}^T \frac{36g_t^2}{\zeta^2}\ln(\frac{2}{\delta})}
\end{equation} 
\end{lemma}

\textit{Proof}
Let $Z_l$ and $Y_t$ be defined as follows:
\begin{align*}
&Z_l = \text{regret}(l) - \frac{3g_l}{\zeta}s_l - \frac{2g_l}{\zeta l^2}s_l\\
& Y_l= \sum_{l=1}^t Z_l
\end{align*}

Hence, $Y_t$ is a super-martingale process:
\begin{align*}
    &\mathbb{E}[Y_t-Y_{t-1}|\filtration_{t-1}]=\mathbb{E}[Z_t] = \mathbb{E}[\text{regret}(t)||\filtration_{t-1}] - \frac{3g_l}{\zeta}s_l - \frac{2g_l}{\zeta l^2}s_l\\
& \mathbb{E}[\text{regret}(t)|\filtration_{t-1}] \leq  \mathbb{E}[\reggap|\filtration_{t-1}] \leq \frac{3g_l}{\zeta}s_l + \frac{2g_l}{\zeta l^2}s_l\\
& \mathbb{E}[Y_t-Y_{t-1}|\filtration_{t-1}]\leq 0
\end{align*}
We now apply Azuma-Hoeffding inequality. We define $Y_0=0$. Note that $|Y_{t} -Y_{t-1} |=|Z_l|$ is bounded by $1+3g_l -2g_l$. Hence, $c = 6g_t$.
Setting $a = \sqrt{2\ln(\frac{2}{\delta})\sum_{t=1}^Tc_t^2}$ in the above inequality, we obtain that with probability $1-\frac{\delta}{2}$ ,

\begin{align}
&Y_t \leq \sqrt{2\ln(\frac{2}{\delta})\sum_{t=1}^T36g_t^2}\\
&\sum_{t=1}^T \Big(\text{regret}(t) - \frac{3g_t}{\zeta}s_t - \frac{2 g_t}{\zeta t^2}s_t\Big)\leq \sqrt{2\ln(\frac{2}{\delta})\sum_{t=1}^T36g_t^2}\\
&\sum_{t=1}^T \Big(\text{regret}(t)\Big)\leq \sum_{t=1}^T \frac{3 g_t}{\zeta}s_t +\sum_{t=1}^T  \frac{2g_t}{\zeta t^2}s_t + \sqrt{2\ln(\frac{2}{\delta})\sum_{t=1}^T36g_t^2}
\end{align}
$\hfill\square$

\begin{lemma}(\textnormal{Azuma-Hoeffding}). If a super-martingale $Y_t$ (with $t\geq 0$) and its the corresponding filtration 
$\filtration_{t-1}$, satisfies  $|Y_t-{Y_{t-1}}|\leq ct$ for some constant $c$ for all $t=1,\dots,T$ then for any $x\geq0$:
\begin{equation}
Pr(Y_t-Y_0\geq x) \leq exp\Big(\frac{-x^2}{2\sum_{t=1}^Tc_t^2} \Big)
\end{equation} 
\end{lemma}

\begin{lemma}
\label{std}
$\sum_{t=1}^T s_{A_t,t}  \leq   \sqrt{dNT}\sqrt{C\Big(\log\Big(\frac{( \trace{\Sigma_\weightvector}+\trace{\Sigma_u}+\trace{ \Sigma_u^{-1})}}{d}+\frac{T}{\sigma_\epsilon^2dN}\Big)\Big)}$
\end{lemma}

For simplicity, we let $s_{A_t,t} = s_t$ below. 
\begin{align*}
\deti{\partA}+\deti{\partB})&=\deti{ \mathbf{1}_{N\times N}}^d\deti{ \Sigma_w}^N+\deti{ \mathbf{I}_{N}}^d\deti{ \Sigma_u}^N\\
&=\deti{ \Sigma_u}^N
\end{align*}

\begin{align*}
\log(\deti{\Omega_t})&\geq \log(\deti{\priorcov})+\sum_{t=1}^T\log(1+\frac{s_t^2}{\sigma_\epsilon^2}) \\
&\geq  \log(\deti{\partA}+\deti{\partB})+\sum_{t=1}^T\log(1+\frac{s_t^2}{\sigma_\epsilon^2})\\
&=n\log(\deti{ \Sigma_u})+\sum_{t=1}^T\log(1+\frac{s_t^2}{\sigma_\epsilon^2})
\end{align*}

\begin{align*}
\trace{\Omega_t}&\leq \trace{\priorcov}+\frac{T}{\sigma_\epsilon^2}\\
&= \trace{\partA}+\trace{\partB}+\frac{T}{\sigma_\epsilon^2}\\
&= \trace{ \mathbf{1}_{N\times N}}\trace{\Sigma_w}+\trace{ \mathbf{I}_{N}}\trace{\Sigma_u}+\frac{T}{\sigma_\epsilon^2}\\
&=  N\trace{\Sigma_w}+N\trace{\Sigma_u}+\frac{T}{\sigma_\epsilon^2}\\
\end{align*}

Using the determinant-trace inequality, we have the following relation:

\begin{align*}
&\Big(\frac{1}{dN}\trace{\Omega_t}\Big)^{dN}\geq\deti{\Omega_t}\\
&dN\log(\frac{1}{dN}\trace{\Omega_t})\geq \log(\deti{\Omega_t})\\
\end{align*}
\begin{align*}
&dN\log(\frac{1}{dN}\trace{\Omega_t})\geq \log(\deti{\Omega_t})\\
&dN\log(\frac{1}{dN}(\trace{\priorcov}+\frac{T}{\sigma_\epsilon^2}))\geq \log(\deti{\Omega_t})\geq N\log(\deti{ \Sigma_u})+\sum_{t=1}^T\log(1+\frac{s_t^2}{\sigma_\epsilon^2})\\
&dN\log(\frac{1}{dN}(\trace{\priorcov}+\frac{T}{\sigma_\epsilon^2}))\geq N\log(\deti{ \Sigma_u})+\sum_{t=1}^T\log(1+\frac{s_t^2}{\sigma_\epsilon^2})\\
&dN\log(\frac{1}{dN}(\trace{\priorcov}+\frac{T}{\sigma_\epsilon^2}))- N\log(\deti{ \Sigma_u})\geq\sum_{t=1}^T\log(1+\frac{s_t^2}{\sigma_\epsilon^2})\\
&dN\log(\frac{1}{dN}(\trace{\priorcov}+\frac{T}{\sigma_\epsilon^2}))+ N\log(\deti{ \Sigma_u^{-1})}\geq\sum_{t=1}^T\log(1+\frac{s_t^2}{\sigma_\epsilon^2})\\
&dN\log(\frac{1}{dN}(\trace{\priorcov}+\frac{T}{\sigma_\epsilon^2}))+ dN\log(\frac{1}{d}\trace{ \Sigma_u^{-1})}\geq\sum_{t=1}^T\log(1+\frac{s_t^2}{\sigma_\epsilon^2})\\
\end{align*}
\begin{align*}
&dN\big(\log(\frac{1}{dN}(\trace{\priorcov}+\frac{T}{\sigma_\epsilon^2}))+ \log(\frac{1}{d}\trace{ \Sigma_u^{-1})}\big)\geq\sum_{t=1}^T\log(1+\frac{s_t^2}{\sigma_\epsilon^2})\\
&dN\big(\log((\frac{\trace{\priorcov}\sigma_\epsilon^2+T}{\sigma_\epsilon^2dN}))+ \log(\frac{1}{d}\trace{ \Sigma_u^{-1})}\big)\geq\sum_{t=1}^T\log(1+\frac{s_t^2}{\sigma_\epsilon^2})\\
&dN\big(\log(\frac{\trace{\priorcov}\sigma_\epsilon^2+T+N\trace{ \Sigma_u^{-1}}\sigma_\epsilon^2}{\sigma_\epsilon^2dN})\big)\geq\sum_{t=1}^T\log(1+\frac{s_t^2}{\sigma_\epsilon^2})\\
&dN\big(\log(\frac{( n\trace{\Sigma_w}+N\trace{\Sigma_u})\sigma_\epsilon^2+T+N\trace{ \Sigma_u^{-1}}\sigma_\epsilon^2}{\sigma_\epsilon^2dN})\big)\geq\sum_{t=1}^T\log(1+\frac{s_t^2}{\sigma_\epsilon^2})\\
&dN\big(\log(\frac{( \trace{\Sigma_w}+\trace{\Sigma_u}+\trace{ \Sigma_u^{-1})}}{d}+\frac{T}{\sigma_\epsilon^2dN})\big)\geq\sum_{t=1}^T\log(1+\frac{s_t^2}{\sigma_\epsilon^2})\\
\end{align*}
Let, $s_t^2 \leq \eminu$. For all $y \in [0,\eminu]$ $\log(1+\frac{y}{\sigma_\epsilon^2})\geq \frac{1}{\eminu}\log(1+\frac{\eminu}{\sigma_\epsilon^2})y$ \\
(See argument in \cite{vaswani2017horde}). 

\begin{align*}
\log(1+\frac{s_t^2}{\sigma^2}) \geq  \frac{1}{\eminu}\log(1+\frac{\eminu}{\sigma_\epsilon^2})s_t^2)\\
\frac{1}{\eminlam\log(1+\frac{1}{\eminlam\sigma_\epsilon^2})} \log(1+\frac{s_t^2}{\sigma_\epsilon^2})\leq s_t^2 
\end{align*}

\begin{align*}
\sum_{t=1}^T s_t^2 \leq C\sum_{t=1}^T\log(1+\frac{s_t^2}{\sigma_\epsilon^2})
\end{align*}

Where, $C= {\eminlam\log(1+\frac{1}{\eminlam\sigma_\epsilon^2})}$ 

By Cauchy Schwartz 
\begin{align*}
\sum_{t=1}^T s_t \leq \sqrt{T}\sqrt{\sum_{t=1}^T s_t^2 }\\
\sum_{t=1}^T s_t \leq \sqrt{T}\sqrt{C\sum_{t=1}^T\log(1+\frac{s_t^2}{\sigma_\epsilon^2}}) 
\end{align*} 

\begin{align*}
\sum_{t=1}^T s_t \leq \sqrt{T}\sqrt{CdN\Big(\log\Big(\frac{( \trace{\Sigma_w}+\trace{\Sigma_u}+\trace{ \Sigma_u^{-1})}}{d}+\frac{T}{\sigma_\epsilon^2dN}\Big)\Big)}\\
\sum_{t=1}^T s_t \leq \sqrt{dNT}\sqrt{C\Big(\log\Big(\frac{( \trace{\Sigma_w}+\trace{\Sigma_u}+\trace{ \Sigma_u^{-1})}}{d}+\frac{T}{\sigma_\epsilon^2dN}\Big)\Big)}
\end{align*}
\hfill$\square$\\

\section{Simulation}
\label{sec:supplement}
We include additional information about the simulation environment.  We first explain general information about the simulation environment. We then provide the procedures for generating state variables (features) in the simulation. Finally, we discuss how we used \HSVone{} to arrive at the feature representations used in the simulation.

\textbf{Simulation dynamics} Within the simulation states are updated every thirty minutes. Each thirty minutes is associated with a date-time, thus we can acquire the month from the current time which is useful in updating the temperature.  The decision times are set roughly two hours apart from 9:00 to 19:00.

\textbf{Availability} In the real-study users are not always available to receive treatment for a suite of reasons. For example, they may be driving a vehicle or they might have recently received treatment. Thus, at each decision time we update the context feature $Available_i \sim Bernoulli(.8)$. for the $i^{th}$ user where $Available_i$ is drawn from a Bernoulli.  This condition reduces the distance between the settings in the environment and those in a real-world study. At each decision time interventions are only sent to users who are available; i.e.  user $i$ cannot receive an intervention when $Available_i =0$. 

\textbf{Recruitment} We follow the recruitment rate observed in $\HSVone$. For example, if 20\% of the total number of participants were recruited in the third week of $\HSVone$ we recruit 20\% of the total number of participants who will be recruited in the third week of the simulation. To explore the effect of running the study for varying lengths we scale the recruitment rates. For example, if the true study ran for 8 weeks, and we want to run a simulation for three weeks, we proportionally scale the recruitment in each of the three weeks so that the relative recruitment in each week remains the same.  
In these experiments we would like to recruit the entire population within 6 weeks. Thus about 10\% of participants are recruited each week, except for the second week of the study where about 30\% of all participants are recruited. This reflects the recruitment rates seen in the study, which were more of less consistent throughout besides one increase in the second week.

We generate states from historical data. Given relevant context we search historical data for states which match this given context. This subset of matching states can be used to generate new states. We discuss this in more detail in \secref{query}. Then, we describe in more detail how we generate temperature, location and step counts. 

\subsection{Querying history}
\label{query}
 \algref{alg:statefunctions} is used to obtain relevant historical data in order to form a probability distribution over some target feature value. 
 For example, if we would like a probability distribution over discretized temperature IDs under a given context, we would search over the historical data for all temperature IDs present under this context. This set of context-specific temperature IDs can then be used to form a distribution to simulate a new ID. This process  of querying historical data is used throughout the simulation and is outlined in \algref{alg:statefunctions}. For example, it is used in generating new step counts, new locations and new temperatures.

 \begin{algorithm}[H]
 \caption{$\statefunctions$}
 \label{alg:statefunctions}
 \begin{algorithmic}[1]
\STATE{\textsc{INPUT} =  historical data $   [\mathbf{x_i} ; i = [1,N] ]$, conditioning state $\mathbf{x^*}$, target data variable $y=f(x)$ }, 
\STATE{$\mathcal{S}= \{\}$}
\FOR{$i=1$ to $N$}
\IF{$\mathbf{x_i}==\mathbf{x^*}$}
\STATE{Add $f(x_i)$ to $\mathcal{S}$}
\ENDIF
\ENDFOR
\STATE{\textsc{OUTPUT} =$\mathcal{S}$}
\end{algorithmic}
 \end{algorithm}

As the simulation environment simulates draws stochastically from a variety of probability distributions, 
it is possible it draws a state which was not present in the historical dataset. In this case there is a process
for finding a matching state. Similarly we might have a state in the historical dataset with insufficient samples 
to form an informative (not overly-noisy) distribution. In this case we also find a surrogate state with which 
to generate future step counts.
The idea of the process is to find the closest state to the current state, such 
that this close state has sufficient data to generate a good distribution. 
Again, given a state, we want to be able to generate a step count from a distribution 
with sufficient data to inform its parameters. The pseudocode for how we do so is shown in \algref{findmatch}

This algorithm takes as input a target state, $s^*$. We also have a dictionary(hasmap) formed from the 
historical dataset. The keys to this dictionary are the states which existed in the dataset. A value is
an array of step counts for this state. 

\begin{algorithm}[H]
	\caption{\textsc{\small{FindMatch}}\label{findmatch}}
	\begin{algorithmic}[1]
		
		\STATE{\textsc{INPUT} =  current state $s^* \in \mathbb{R}^d$, dictionary of existing states to step counts $\mathbb{D}=\{s:[c_1,\dots,c_N]\}$} 
		\STATE{match$\leftarrow$None}
		\IF{$s^* \in \mathbb{D}$ and $len( \mathbb{D}[s^*])>30$}
		\STATE{match$\leftarrow s^*$}
		\ELSE
		\STATE{$new\_size$ = d-1}
		\WHILE{match is None}
		\STATE{\#find state of size new size with most data points in historical dataset}
		\STATE{form new states of size $new\_size$}
		\STATE{rank states $s$ by $len( \mathbb{D}[s])$}
		\STATE{choose state with greatest len}
		\STATE{$temp\leftarrow max_s len( \mathbb{D}[s])$}
		\IF{$ \mathbb{D}[temp] > 30$}
		\STATE{match$\leftarrow temp$}
		\ENDIF
		\STATE{$new\_size=new\_size-1$}
		\ENDWHILE
		\ENDIF
	\end{algorithmic}
\end{algorithm}

This procedure gives the closest state with the most data points to our current state.

To be more explicit about lines 8-11. A state is a vector of some length, for example $[1,0,1]$. 
When we consider all subsets of size 2, we are considering the subsets $[1,0]$,$[1,1]$, and $[0,1]$. 
For each of these we can look in the historical data set and find all points where this state was true. 
Thus for each subset we'll get a new list of points,  $[1,0] = [c_1,\dots,c_{N1}]$ $[1,1] = [c_1,\dots,c_{N2}]$,
$[0,1] = [c_1,\dots,c_{N3}]$. We now look at $N1,N2,N3$ and choose the state with the highest value. For example, 
if the lists were:  $[1,0] = [c_1,\dots,c_{100}]$ $[1,1] = [c_1,\dots,c_{2}]$,
$[0,1] = [c_1,\dots,c_{300}]$, we would choose $s=[0,1]$.
Now if we encounter the state $[1,0,1]$ and there is insufficient data to form a distribution from this state, 
we will instead form it from the values found under the state $[0,1]$, $[c_1,\dots,c_{300}]$.

\subsection{Generating temperature}
We mimic a trial where 
everyone resides in the same general area, such as a city. In this setting everyone experiences the same global temperature. 
We describe how to obtain temperature at any point in time in \algref{get_temp}. The temperature is updated exactly five times a day.

In the following algorithms $ \simutime,$ refers to a timestamp, $\history$ refers to a historical dataset, $K_t$ refers to a set of temperature IDs, and $\weather_{\simutime-1}$ refers to the temperature at the previous time stamp. Here,  $\history=\HSVone$ and $K_t = \{\text{hot},\text{cold}\}$. The contextual features which influence temperature are time of day, day of the week and the month $tod$, $dow$ and $month$ respectively. Furthermore, at all times besides the first moment in the trial, the next temperature depends on the current temperature $\weather_{\simutime-1}$.

\begin{algorithm}[H]
	\caption{\textsc{\small{GetTemperature}}\label{get_temp}}
	\begin{algorithmic}[1]
		
		\STATE{$\textsc{INPUT} =  \simutime,\history,K_t,\weather_{\simutime-1},$} 
		 \STATE{$tod\leftarrow tod(\simutime)$}
 		\STATE{$dow\leftarrow dow(\simutime)$}
 		\STATE{$month\leftarrow month(\simutime)$}
		\IF{$\weather_{\simutime-1}$ is Null}
		\STATE{$q \leftarrow [tod,dow,month]$}
		\ELSE
		\STATE{$q \leftarrow [tod,dow,month,\weather_{\simutime-1}]$}
		\ENDIF
		\STATE{$\textbf{p}\leftarrow [0] _{K_l}$}
		\STATE{$\mathcal{T}\leftarrow \statefunctions(\history,q,\weather)$}
		\FOR{$k \in K_t$}
		\STATE{$p_k = \frac{1}{|\mathcal{T}|}\sum_{i=0}^{|\mathcal{T}|}\mathbbm{1}_{l_i==k}$}
		\ENDFOR
		\STATE{$\weather_\simutime \sim Categorical([p_{cold},p_{hot}])$}
		\STATE{\textsc{OUTPUT} $\weather_\simutime$ }	
	\end{algorithmic}
\end{algorithm}

\subsection{Generating location}
In the following algorithms $ \simutime,$ refers to a timestamp, $\groupid_\user$ refers to the group id of user $i$,$\history$ refers to a historical dataset, $K_t$ refers to a set of location IDs, and $\l_{\simutime-1}$ refers to the location at the previous time stamp. Here,  $\history=\HSVone$ and $K_t = \{\text{other},\text{home or work}\}$. 

As in generating temperature,  
the contextual features which influence location are time of day, day of the week and the month $tod$, $dow$ and $month$ respectively. 
Generating location is different from generating temperature in that each user moves from location to location independently. Whereas we 
model users to share one common temperature, they move from one location to another independently of other users. Thus we also include 
group id in determining the next location for a given user.

\begin{algorithm}[H]
	\caption{\textsc{\small{GetLocation}}}
	\begin{algorithmic}[1]
		
		\STATE{$\textsc{INPUT} =  \simutime,\groupid_\user,\history,K_l$} 
		 \STATE{$tod\leftarrow tod(\simutime)$}
 		\STATE{$dow\leftarrow dow(\simutime)$}
		\STATE{Find $\starttime$ in $\history$}
		\IF{$\location_{\simutime-1}$ is Null}
		\STATE{$q \leftarrow [tod,dow,\groupid_\user]$}
		\ELSE
		\STATE{$q \leftarrow [tod,dow,\groupid_\user,\location_{\simutime-1}]$}
		\ENDIF
		\STATE{$\mathcal{L}\leftarrow \statefunctions(\history,q,\location)$}
		\STATE{$\textbf{p}\leftarrow [0] _{K_l}$}
		\FOR{$k \in K_l$}
		\STATE{$p_k = \frac{1}{|\mathcal{L}|}\sum_{i=0}^{|\mathcal{L}|}\mathbbm{1}_{l_i==k}$}
		\ENDFOR
		\STATE{$\location_\simutime \sim Categorical([p_{\text{other}},p_{\text{home or work}}])$}
		\STATE{\textsc{OUTPUT} $\location_\simutime$ }
		
	\end{algorithmic}
\end{algorithm}

\subsection{Generating step-counts}
A new step-count is generated for each \patient{} active in the study, every thirty-minutes  according to one of the following scenarios:
\begin{figure}
\begin{enumerate}
\item \patient{} is at a decision time
\begin{enumerate}
\item \patient{} is available
\item \patient{} is not available
\end{enumerate}
\item \patient{} is not at a decision time
\end{enumerate}
\end{figure}

Scenarios 1b and 2 are equivalent with respect to how step-counts are generated; a \patient's step count either depends on whether or not they received an intervention (when they are at a decision time and available) or it does not (because they were either not at a decision time or not available).
Recall, that if a user is available the final step count is generated according to \eqnref{steps}.This equation requires sufficient statistics from \HSVone{}.  The procedure for obtaining these statistics is shown explicitly in \algref{get_steps}.

  \begin{equation}\label{steps}R_{i,k} = \mathbf{N}(\mu_{h(\state_{i,k})},\sigma^2_{h(\state_{i,k})})+ A_{i,k}( f(\state_{i,k})^T\beta_{i} + Z_i). 
  \end{equation}

 \begin{algorithm}[H]
 \caption{\textsc{StepStatistics} \label{get_steps}}
 \label{alg:cohort}
 \begin{algorithmic}[1]
 \STATE{\textsc{INPUT}  $\text{=} \simutime,\groupid^\user,\weather_\simutime,\user,\history$}
 \STATE{\#Compute variables included in conditioning context}
 \STATE{$tod\leftarrow tod(\simutime)$}
 \STATE{$dow\leftarrow dow(\simutime)$}
 \STATE{$y\leftarrow yst(\simutime,\user)$}
\STATE{$q \leftarrow [\groupid^\user,\weather_\simutime,tod,dow,y,\location_{\simutime,\user},\simaction]$}
\STATE{\#Obtain step counts from $\history$ conditioned on $q$}
\STATE{$\mathcal{S}\leftarrow \statefunctions(\history,q,\stepcount)$}
 \STATE{$\hat{\mu}_{\mathcal{S}} \leftarrow \frac{1}{|\mathcal{S}|}\sum_{i=0}^{|\mathcal{S}|}s_i$}
  \STATE{$\hat{\sigma}^2_{\mathcal{S}} \leftarrow \frac{1}{|\mathcal{S}|}\sum_{i=0}^{|\mathcal{S}|}(s_i-\hat{\mu}_{\mathcal{S}})^2$}
   \STATE{\textsc{OUTPUT}  $\hat{\mu}_{\mathcal{S}},\hat{\sigma}^2_{\mathcal{S}} $}
\end{algorithmic}
 \end{algorithm}
 Here, $\simutime,\groupid^\user,\weather_\simutime,\location_\user,\history$ refer to the current time in the trial, the group id of the $i^{th}$ user, the temperature at time $t$, the location of the $i^{th}$ user, and a historical dataset, respectively. To find sufficient statistics of step counts, we also employ the time of day and day of the week,  $tod$ and $dow$ respectively. 
Finally, $yst(\simutime,\user)$ describes the previous step count as high or low.

\paragraph{}{Settings for \populationgen{}}
\begin{table}
\centering	
\resizebox{\columnwidth}{!}{%
\begin{tabular}{|p{1.5cm |}|p{1.5cm |}|p{1.5cm |}||}
\hline
   \multicolumn{1}{|c|}{Homogeneous}&\multicolumn{1}{c|} {Bi-modal } & \multicolumn{1}{c|}{Smooth} \\
\hline
    \multicolumn{1}{|c|}{$Z^i =0$ $\beta^l_i$=0} &   \multicolumn{1}{r|}{ $ Z_i,\beta^l_i = \begin{cases}
      0.1, 0.l & \text{if}\ i \in \text{group one} \\
      -0.3,-0.l &  \text{if}\ i \in  \text{group two}
    \end{cases}$ } &  \multicolumn{1}{r|}{ $Z_i \sim \mathcal{N}(0,0.35)$ $\beta^l_i\sim \mathcal{N}(0,0.1)$} \\
    \hline
\end{tabular}
}
\caption{\textmd{Settings for Z in three cases of homogeneous, bimodal and smoothly varying populations. } \label{table:Z}}
\end{table}

\section{Feature construction}
\label{sup:features}
We provide more details on the processes used for feature construction. As stated in the paper we rely heavily on the dataset  \HSVone{} to make all feature construction decisions. The one exception is in the design of the location feature, for which we had domain knowledge to rely on (more detail below)

\subsection{Baseline activity} Each user is assigned to one of two groups: a low-activity group or a high-activity group. These groups are found from the historical data. We perform hierarchical clustering using the method \textit{hcluster} in  scikit-learn \cite{scikit-learn}. We used a euclidean distance metric to cluster the data and found that two groups naturally arose. These groups were consistent with the population of  \HSVone{}, which consisted of participants who were generally either office administrators or students. 

\subsection{State features}
We now briefly outline the decisions for the remaining features: time of day, day of the week, and temperature. For each feature we explored various categorical representations. For each, 
the question was how many categories to use to represent the data. For each feature we followed the same procedure. 

\begin{enumerate}
\item We chose a number of categories ($k$) to threshold the data into
\item We partitioned the data into $k$ categories
\item We clustered the step counts according to these $k$ categories
\item We computed the Calinski-Harabasz score of this clustering
\item We chose the final $k$ to be that which provided the highest score 
\end{enumerate}

For example, consider the task of representing temperature. Let $l$ be a temperature, $x$ be a step count and $x_{l_b}$ be a thirty-minute step count occurring when the temperature $l$ was assigned to bucket $b$. Given a historical dataset, we have a vector $\bf{x}$ where each entry $x_{i,t}$ refers to the  thirty-minute step count of user $i$ at time $t$.

\begin{itemize}

\item Let $p$ be a number of buckets. We create $p$ buckets by finding quantiles of $l$. For example, if $p$=2, we find the $50^{th}$ quantile of $l$. A bucket is defined by a tuple of thresholds $(th_1,th_2)$, such that for a data point $d$ to belong to bucket $i$, $d$ must be in the range of the tuple ($th_1\leq d< th_2$).

\item For each temperature, we determine the bucket label which best describes this temperature. That is the label $y$ of $l$, is the bucket for which $th^y_1\leq \bar{s}^l< th^y_2$.

\item We now create a vector of labels $y$, of the same length as $\bf{x}$. Each  $y^l_{i,t}$ is the bucket assigned to $l_{i,t}$. For example, if the temperature for user $i$ at time  $t$ falls into the lowest bucket, $0$ would be the label assigned to $l_{i,t}$. This induces a clustering of step-counts where the label is a temperature bucket. 

\item We determine the Calabrinski-Harabasz score of this clustering. 

\end{itemize}

We test this procedure from $p$ equal to 1, through 4. 

For example, consider determining a representation for time of day. We choose a partition to be morning, afternoon, evening. For each thirty-minute step count, if it occurred in the morning we assign it to the morning cluster, if it occurred in the afternoon we assign it to the afternoon cluster, etc. Now we have three clusters of step counts and we can compute the C score of this clustering. We repeat the process for different partitions of the day. 

\textbf{Time of day} To discover the representation for time of day which best explained the observed step counts, we considered all sequential partitions from length 2-8. We found that early-day, late-day, and night best explained the data. 

\textbf{Day of the week} To discover the representation for day of the week which best explained the observed step counts, we considered two partitions: every day, or weekday/weekend. We found weekday/weekend to be a better fit to the data. 

\textbf{Temperature} Here we choose different percentiles to partition the data. We consider between 2 and 5 partitions (percentiles at 50, to 20,40,60,80). Here we found two partitions to best fit the step counts. We also tried more complicated representations of weather combined with temperature, however for the purpose of this paper we found a simple representation to best allow us to explore the relevant questions in this problem setting. 

\textbf{Location} In representing location we relied on domain knowledge. We found that participants tend to be more responsive when they are either at home or work, than in other places. Thus, we decided to represent location as belonging to one of two categories: home/work or other.

\section{Feasibility Study}
\label{feature_clinical}
In the clinical trial we describe users' states with the features described in Table 4. The two  features which differ from the simulation environment are engagement and exposure to treatment. We clarify these features below. 

\textbf{Engagement}
The engagement variable measures the extent to which a user engages with the mHealth application deployed in the trial. There are several screens within the application that a user can view. Across all users we measure the $40^{th}$ percentile of number of screens viewed on day $d$. If user $i$ views more than this percentile, we set their engagement level to 1, otherwise it is 0.

\textbf{Exposure to treatment}
This variable captures the extent to which a user is treated, or the treatment dosage experienced by this user. Let $D_i$ denote the exposure to treatment  for user $i$. Whenever a message is delivered to a user's phone $D_i$i s updated. That is, if a message is delivered between time $t$ and $t+1$, $D_{t+1}=\lambda D_t+1$. If a message is not delivered, $D_{t+1}=\lambda D_t$. Here, we se $\lambda$ according to data from \HSVone{} and initialize $D$ to 0.

\end{document}